%% file: main.tex
\documentclass{article}

\PassOptionsToPackage{numbers,sort&compress}{natbib}

\usepackage[position,final]{neurips_2026}  

\usepackage[show]{chato-notes}
\usepackage[utf8]{inputenc}
\usepackage[T1]{fontenc}
\usepackage{microtype}
\usepackage{graphicx}
\usepackage{subcaption}
\usepackage{tikz}
\usepackage{forest}
\usepackage{adjustbox}
\usetikzlibrary{backgrounds, positioning,arrows.meta,shadows.blur,fit}
\usetikzlibrary{shadows}
\usepackage{booktabs}
\usepackage{amsthm}
\usepackage{hyperref}
\usepackage{url}
\usepackage[most]{tcolorbox}
\usepackage{amsmath}
\usepackage{amssymb}
\usepackage{mathtools}
\usepackage[capitalize,noabbrev]{cleveref}
\usepackage[textsize=tiny]{todonotes}
\usepackage{xcolor}

\definecolor{typeModel}{RGB}{240, 240, 240}      
\definecolor{typeExplainer}{RGB}{220, 235, 255}  
\definecolor{typeExplanation}{RGB}{225, 245, 225}
\definecolor{typeBehavior}{RGB}{255, 230, 230}   
\definecolor{typeIntervention}{RGB}{255, 250, 210}

\definecolor{boxDelta}{RGB}{230, 240, 255}
\definecolor{boxXAI}{RGB}{255, 235, 205}

\tikzset{
    block/.style={
        draw, rectangle, rounded corners,
        minimum width=2.4cm, minimum height=1.1cm,
        align=center, fill=white,
        inner sep=4pt, font=\small
    },
    macrobox/.style={
        draw, dotted, thick, rounded corners,
        fill=deltablue, inner sep=20pt
    },
    xabox/.style={
        draw, solid, rounded corners,
        fill=xaorange, inner sep=12pt
    },
    explained/.style={
        dashed, <->, >=stealth
    }
}

\newcommand{\rev}[1]{{#1}}

\theoremstyle{plain}

\theoremstyle{definition}

\theoremstyle{remark}

\newtheorem{exampleinner}{Example}
\newenvironment{example}{%
  \begin{tcolorbox}[
    colback=gray!8,
    colframe=gray!8,
    left=4pt, right=4pt, top=2pt, bottom=2pt,
    boxrule=0pt,
    arc=2pt,
    boxsep=0pt,
    before skip=4pt,
    after skip=4pt
  ]
  \begin{exampleinner}
}{%
  \end{exampleinner}
  \end{tcolorbox}
}

\input{defines}

\title{Comparing Explanations is Not Enough, Explain the Change: New Standards are Needed to Explain Behavioral Shifts in Large Language Models}

\author{%
  Martino Ciaperoni\thanks{Equal contribution} \\
  Scuola Normale Superiore \\
  Pisa, Italy \\
  {\scriptsize\texttt{martino.ciaperoni@sns.it}}
  \And
  Marzio Di Vece$^*$ \\
  Scuola Normale Superiore \\
  Pisa, Italy \\
  {\scriptsize\texttt{marzio.divece@sns.it}}
  \And
  Roberto Pellungrini \\
  Scuola Normale Superiore \\
  Pisa, Italy \\
  {\scriptsize\texttt{roberto.pellungrini@sns.it}}
  \AND
  Luca Pappalardo \\
  ISTI-CNR \\
  Pisa, Italy \\
  {\scriptsize\texttt{luca.pappalardo@isti.cnr.it}}
  \And
  Fosca Giannotti \\
  Scuola Normale Superiore \\
  Pisa, Italy \\
  {\scriptsize\texttt{fosca.giannotti@sns.it}}
  \And
  Francesco Giannini \\
  University of Pisa \\
  Pisa, Italy \\
  {\scriptsize\texttt{francesco.giannini@unipi.it}}
}

\begin{document}

\maketitle

\begin{abstract}
Large-scale foundation models exhibit \emph{behavioral shifts} when subjected to interventions such as scaling, fine-tuning, reinforcement learning with human feedback, or in-context learning. Current explainability methods are structurally ill-suited to explain these shifts, because  they either treat models as static objects, as traditional eXplainable AI (XAI) approaches do, or merely compare independent explanations across different checkpoints of a model. As a result, these approaches fail to explain the functional transition between two model instances in which a certain behavior has shifted following an intervention. This gap creates significant governance risks across jurisdictions including the EU AI Act, US state legislation, and Chinese AI regulations, which require documenting causal chains for substantial system modifications. 
This position paper argues that explaining behavioral shifts in large language models requires a principled approach that treats the shift itself as the primary object of explanation: namely, one that explains how and why an intervention transforms a reference model into an updated model with different behavior.
To support this claim, we introduce \textit{Comparative} XAI (\acr{}), a novel XAI paradigm aimed at explaining the difference between two model checkpoints where a behavior has shifted, together with a set of desiderata specifying what \acr{} explainers and explanations must satisfy, including comparability, validity, actionability, and monitoring, with the goal of grounding model auditing in explicit, measurable requirements.
Finally, we provide preliminary evidence suggesting the need for \acr{} in practice through illustrative experiments, compiling the resulting findings into a transition report directly usable for governance and incident documentation.
\end{abstract}

\section{Introduction}\label{sec:intro}

Large-scale foundation models, including Large Language Models (LLMs), exhibit various observable qualitative and quantitative properties, such as alignment to human values, sycophancy, or performance on specific tasks, which we broadly refer to as \textit{behaviors}. Recent studies have revealed that these models are prone to striking \textit{behavioral shifts}: systematic and often unpredictable changes in these behaviors across model updates, prompts, or evaluation settings that were not explicitly programmed~\citep{berti2025emergent}.
Behavioral shifts can be both beneficial and detrimental.
On the positive side, LLMs may acquire useful general abilities, such as in-context learning~\citep{brown2020language} and tool use through API calls~\citep{schick2023toolformer}.
On the negative side, these models may develop deceptive and manipulative behaviors: GPT-4 is able to persuade a human to solve a CAPTCHA by feigning visual impairment~\citep{openai2023gpt4}, while Claude can engage in reward hacking in a coding task, with the behavior generalizing into lying and sabotage~\citep{macdiarmid2025misalignment}.
Some behavioral shifts do not arise gradually but manifest abruptly. A growing body of literature focuses on \emph{emergent abilities}, i.e., 
sudden gains in task performance~\citep{wei2022} driven by factors such as  changes in scale~\citep{Du2024}, training procedures~\citep{Lu2024,snell2024}, or evaluation choices~\citep{mirage}, and 
\textit{emergent misalignment}, i.e., unexpected onset of undesirable behaviors~\citep{DBLP:conf/icml/BetleyTWSBSLE25,DBLP:journals/corr/abs-2506-11613}.
However, much of the existing literature remains primarily descriptive: it documents \emph{what} behavioral shifts occur and under which conditions, but offers limited insight into \emph{why} they arise. 
Because LLM input spaces are practically unbounded, finite benchmarks expose only behavioral symptoms, not root causes. Without isolating the origins of these shifts, targeted mitigation fails, reducing model safety alignment and control to mere trial-and-error.
There is therefore a need to move beyond behavioral observation and explicitly explain the internal mechanisms driving behavioral shifts. 

\textbf{Single-checkpoint XAI cannot explain shifts.} To explain behavioral shifts, one might intuitively turn to the vast ecosystem of traditional (i.e., single-checkpoint) eXplainable AI (XAI) tools that have been introduced over the years~\citep{guidotti2018survey,calderon2025behalf,zhao2024explainability,ciaperoni2026virgil}, including feature attribution methods~\citep{DBLP:conf/nips/LundbergL17,DBLP:conf/naacl/Ribeiro0G16,integratedgradients}, probing~\citep{belinkov2022}, concept-based methods~\citep{kim2018interpretability}, and mechanistic interpretability~\citep{conmy2023towards,zhang2025controlling,gantla2025exploring}. 
However, these methods are fundamentally designed to answer \emph{"why does this model produce this  output?"} not \emph{"what changed between two model versions?"} When applied to an updated model obtained through a certain intervention, a single-checkpoint explainer can identify which features drive current outputs, but it suffers from explanatory underdetermination: it cannot determine whether those features were already present in a previous version of the model used as a reference, or if they emerged strictly as a consequence of the intervention. 
This underdetermination is not a limitation of a particular method: it is a structural consequence of operating on a single model checkpoint without a reference.
For instance, applying standard feature attribution methods to a fine-tuned model alone can identify salient tokens for prediction, but cannot reveal \textit{which} attributions shifted due to fine-tuning and which were already present in the base model. Even more, these methods cannot explain \textit{why}  certain attributes have changed due to the intervention on the reference model.

\textbf{Emerging comparative approaches are still unsuitable.}
Motivated by the limitations of single-checkpoint methods, a growing but fragmented body of work has begun comparing model checkpoints directly.
For instance, recent works have contrasted pretrained and fine-tuned checkpoints via layerwise similarity measures such as CKA~\citep{kornblith2019similarity,phang2021fine,yao2025pre},
activation  patching~\citep{prakash2024finetuning,soligo2025convergent},
or sparse autoencoders (SAEs) and crosscoders~\citep{cunningham2024sparse,lindsey2024crosscoders,jiralerspong2025crossarch,bricken2024stagediffing,wang2025persona,chen2025persona}.
Beyond model fine-tuning, comparative explanations have also been introduced for continual learning and model drift settings~\citep{nguyen2020dissecting,artelt2023contrasting}.
While comparing explanations across checkpoints is a natural first step, a comparative explanation should satisfy additional requirements to constitute a full and principled account of a behavioral shift. 
Existing approaches are method-specific and aim to address different issues, while falling short of providing an exhaustive solution: representation similarity measures can indicate where models diverge ~\citep{phang2021fine,yao2025pre} and can be stable, but not causally relevant; 
SAEs and crosscoders tell us which latent features have been modified~\citep{lindsey2024crosscoders,jiralerspong2025crossarch,bricken2024stagediffing,wang2025persona,chen2025persona}
leading to a difference in features which may be causally relevant, but not intervention-specific; activation patching and steering~\citep{prakash2024finetuning,soligo2025convergent} tell us which causal mechanisms are involved, but do not provide a clear and intelligible explanation for auditors.
 
\textbf{Towards a principled framework to explain behavioral shifts.} Current attempts highlight that there is a lack of a principled framework specifying \emph{what} a comparative explanation must actually establish, \emph{how robustly} it must be supported, \emph{whether} a detected shift is specific to an intervention rather than generic drift, and \emph{how} these claims should be standardized for oversight.
This absence is a regulatory liability. Emerging AI regulations across jurisdictions now converge on a strict requirement: when an AI system at systemic- or high-risk~\footnote{High-risk systems are those capable of impacting the safety, health, or fundamental
rights of affected persons~\citep{EUAIAct2024, ColoradoAIAct}; systemic-risk systems are
general-purpose "frontier" models with high capacity to impact markets and society~\citep{EC_GPAI_Safety_2025,
CaliforniaSB53}.} undergoes a substantial modification, providers must produce auditable evidence of what changed and rigorously document the causal chain of events leading to any serious incident. The EU AI Act operationalizes this through mandatory risk management logs, incident notification with causal evidence, and renewed conformity assessments for any change altering a system's risk profile~\citep{EUAIAct2024}. Analogous transparency and impact assessments are required by US state legislation~\citep{CaliforniaSB53,ColoradoAIAct} and Chinese algorithmic regulations~\citep{ChinaGenAI2023,ChinaAlgoRegs2022} following significant interventions to the model.
Fulfilling the requirements of modern transition reports~\citep{Juliussen2025,hacker2025finetuning,novelli2024governance} demands a structured, auditable paradigm that moves beyond mere single-checkpoint XAI or differences in explanations. The community needs new standards capable of documenting \emph{what changed}, \emph{when} it occurred, \emph{why} it produced a specific outcome, and \emph{which interventions} can systematically reverse it.

Motivated by the limitations of current solutions, \textbf{we posit that explaining behavioral shifts in large language models in view of governance demands requires the development of a novel paradigm for XAI, i.e. \emph{Comparative} XAI (\acrb{}), where
\acrb{} methods should treat the behavioral shift itself as the primary object of explanation.} 

To lay the foundation for \acr{}, the remainder of this paper is structured as follows. \Cref{sec:framework} introduces the \acr{} framework for explaining behavioral shifts in LLMs. \Cref{sec:desiderata} articulates desiderata that \acr{}  methods must satisfy, and \Cref{sec:illustrative_experiment}  showcases  \acr{} in a concrete illustrative scenario. Finally, \Cref{sec:altviews} discusses alternative views, and \Cref{sec:conclusion} presents conclusions.

\section{The \acr{} Paradigm}
\label{sec:framework}

We define a set of possible interventions $\mathcal{I}$, where $I\in\mathcal{I}$ is a function from one model to its subsequent version (we call each version a \emph{checkpoint}). 
We then consider a \emph{checkpoint sequence} $\mathbf{M}{=}(M_0,\ldots,M_T)$ with $M_t:\mathcal{X} \to \mathcal{Y}$ for some input/output spaces $\mathcal{X},\mathcal{Y}$, generated by successive interventions $I_t\in\mathcal{I}$ such that $M_t {=} I_t(M_{t-1})$ for $t=1,\ldots,T$.

\begin{example}
\label{ex:model_sequence}
Consider a checkpoint sequence $\mathbf{M} {=} (M_0, M_1, M_2)$, where 
$M_0$ is a pre-trained LLM intended to support users asking for medical advice.
A task-specific fine-tuning $I_1 \in \mathcal{I}$ aimed at improving factual knowledge in medicine produces a new checkpoint $M_1 {=} I_1(M_0)$. Then, prompt conditioning $I_2 \in \mathcal{I}$ to elicit step-by-step reasoning produces a checkpoint $M_2 = I_2(M_1)$. 
\emph{How can we determine whether behavioral shifts occur between $M_0$ and $M_1$, or between $M_1$ and $M_2$?}
\end{example}
Let $b$ denote a \emph{behavior}, i.e. a qualitative or quantitative property of a model or its outputs that can be operationalized with a measurable metric, allowing a precise determination of when $b$ is present, absent, or shifting. Examples of behaviors include sycophancy (measured by sycophancy rate), and safety (measured by safety violation counts or refusal rates), as well as task accuracy on benchmarks such as HellaSwag~\cite{zellers2019hellaswag} for common-sense reasoning, or hallucination tendency measured by FactScore~\cite{min2023factscore}
 or the semantic entropy~\cite{farquhar2024detecting}.
To measure the shift of $b$ within a checkpoint sequence $\mathbf{M}$, we assume to have available a \textit{behavioral metric} $B$
that associates to each checkpoint $M_t$ an evaluation over $b$ (e.g., $B(M_t)\in\mathbb{R}$). 
Examples of concrete behavioral metrics include task accuracy, safety violation counts, refusal rates, deception indicators, or human preference scores.
A \emph{behavioral shift} occurs when the change in a model’s behavior exceeds a threshold $\varepsilon_B$, which depends on the chosen metric $B$ and the application context. 
Formally, for $\mathbf{M}$ at $0 < \bar{t} \leq T$, we define a \textit{behavioral shift} with respect to behavior $b$ if $\|\Delta B(M_{\bar{t}})\|> \varepsilon_B$, where $\Delta B(M_{\bar{t}}) = B(M_{\bar{t}}) - B(M_{\bar{t}-1})$. 
If needed, standard methods for change-point detection~\cite{truong2020selective} can be used to automatically identify candidate values of $\bar{t}$ at which $B(M_t)$ exhibits a statistically significant behavioral shift along the checkpoint sequence.
When a behavioral shift occurs at $\bar{t}$, we denote with \Mpre${=}M_{\bar{t}-1}$ the \emph{reference} model (before the shift) and with \Mpost${=}M_{\bar{t}}$ the \emph{updated} model (after the shift). 
Note that our formulation allows model behavior to be evaluated both globally at a given checkpoint, via $B(M_t)$, or locally for a specific input $x\in\mathcal{X}$, via $B(M_t, x)$. 
\begin{example}
\label{ex:behavior_shift}
Continuing from \Cref{ex:model_sequence}, consider the behavior $b$ of suggesting that the user calls for immediate medical assistance when clinically appropriate.
Let $X_E \subseteq \mathcal{X}$ denote a set of user prompts that describe urgent clinical scenarios and request medical advice to the LLM. 
Given a checkpoint $M_t$ and a prompt $x \in X_E$, we define a binary metric~$B(M_t,x)\in\{0,1\}$ indicating whether or not $M_t(x)$ recommends the user to call for immediate medical assistance. Accordingly, for the set of prompts $X_E$, $B(M_t)$ is  defined as the percentage of prompts for which $B(M_t,x)=1$.    
Models $M_0, M_1, M_2 \in \mathbf{M}$ correctly recommend urgently calling for medical assistance $85\%$, $90\%$ and $20\%$ of the time, respectively. 
With $\varepsilon_B = 10$ percentage points, the observed drop in performance from $M_1$ to $M_2$ exceeds the predefined threshold for non-negligible change, and thus signals the presence of a behavioral shift.
\emph{How can this behavioral shift be explained?}
\end{example}


\textbf{Explaining behavioral shifts.}
In traditional XAI, an \emph{explainer} produces some \emph{explanation} describing a model's decision process.
For example, this explanation may be a saliency map over inputs, a set of activated neurons, a latent representation, a learned symbolic concept, or a logic rule.
Formally, we define an explainer $\Phi^{(b)}$ for a behavior $b$ as a mapping from a model $M$, and possibly an input $x\in \mathcal{X}$, to an explanation $e^{(b)}$, i.e., $\Phi^{(b)}:M \mapsto e^{(b)}$ for global explanations or $\Phi^{(b)}:(M,x) \mapsto e^{(b)}(x)$ for local explanations. 
Single-checkpoint XAI can explain either \Mpre or \Mpost in isolation, but not the intervention nor the behavioral shifts associated with the transition between them. 
On the contrary, \acr{} considers $\Delta B(\Mpost)$ as the explanatory target, possibly contextualizing the explanation to the intervention $I$, where $\Mpost=I(\Mpre)$. 
Therefore, we argue that a different paradigm should be followed. 
In particular, we need \emph{comparative explainers} (or \acr{} \emph{explainers}) that should map multiple checkpoints obtained through certain interventions into \emph{comparative explanations} focusing on the change itself. 
Formally, a comparative explainer $\Phi^{(b)}_\Delta$ extends an XAI method $\Phi$ to produce comparative explanations $e^{(b)}_\Delta$ mapping \Mpre, \Mpost and $I$ into an explanation: 
$e^{(b)}_\Delta := \Phi^{(b)}_\Delta\left( \Mpre, \Mpost, I \right)$.  Figure~\ref{fig:cxai_framework} summarizes how the two paradigms relate.

\begin{figure*}[t!]
    \centering
    \input{diagrams/xaiAndcxai_NIPS_2}
\caption{\textbf{Comparison between classical XAI and the \acr{} framework.}
An XAI explainer produces independent explanations for single checkpoints. In \acr{}, each checkpoint is associated with a behavior measured by a metric $B$, and a behavioral shift occurs when $\Delta B$ exceeds a task-dependent threshold, thereby defining $\Mpre$ and $\Mpost$. An explainer $\Phi^{(b)}$ is applied to each checkpoint, while a comparative explainer $\Phi^{(b)}_\Delta$ produces a shift-focused explanation $e^{(b)}_\Delta$. 
Solid blue edges denote inputs to $\Phi^{(b)}_{\Delta}$, while dashed lines denote that $e^{(b)}_{\Delta}$ is produced to explain a behavioral shift $b$ measured by $B$.
}
\label{fig:cxai_framework}
\end{figure*}

\begin{example}
\label{ex:delta_explanation}
\Cref{ex:behavior_shift} identifies a behavioral shift in the tendency of the LLM to recommend immediate medical assistance.
To explain this shift, let $\Phi^{(b)}$ be a local feature-attribution explainer such as Integrated Gradients~\cite{integratedgradients} that, for a prompt $x \in  X_E$ describing an urgent clinical scenario and a target response recommending immediate medical assistance, produces saliency scores over input tokens.
Applying $\Phi^{(b)}$ to the same prompt $x$ 
yields two explanations
$e^{(b)}_{\mathrm{pre}}(x) {=} \Phi^{(b)}(\Mpre, x)$ and
$e^{(b)}_{\mathrm{post}}(x) {=} \Phi^{(b)}(\Mpost, x)$.
In a traditional XAI setting, each explanation would be interpreted in isolation. For example, inspecting $e^{(b)}_{\mathrm{post}}$ might reveal that tokens associated with symptom minimization or delay (e.g., “mild”, “wait”) receive high attribution when $\Mpost$ discourages urgent care. However, such an explanation does not account for why $\Mpost$ behaves differently from $\Mpre$.
Instead, a comparative explainer $\Phi^{(b)}_\Delta$ 
yields a comparative explanation
$e^{(b)}_\Delta(x) {=} \Phi^{(b)}_\Delta(\Mpre , \Mpost, I)$,
highlighting how attribution mass changes across checkpoints.
For instance, $e^{(b)}_\Delta$ may reveal that, relative to $\Mpre$, $\Mpost$ assigns greater importance to symptom-minimizing descriptors and less importance to urgency cues (e.g., “chest pain”, “shortness of breath”).
\end{example}


\subsection{Discussion}

The simplest instantiation of the \acr{} paradigm involves an endpoint-comparison strategy, where independent explanations, such as feature attribution scores or concept-level activations from sparse autoencoders, are contrasted to highlight shifted influences across model checkpoints. However, \acr{} extends beyond mere differencing to encompass interventional methodologies, including cross-checkpoint activation patching or steering along contrastive representations, which directly test whether a discovered internal difference serves as an actionable lever for the behavioral shift. 
While \acr{} explainers inherit the model-agnostic or model-specific characteristics of traditional XAI, they are further distinguished by being intervention-agnostic or intervention-specific. For example, an \acr{} method tailored to in-context learning can attribute behavioral changes directly to the tokens introduced in the prompt, identifying the specific causal drivers of the transition. 
Analogously to traditional XAI, \acr{} is also compatible with varying levels of model access: white-box
access unlocks the full range of probing, representation similarity, and mechanistic interventions; black-box or logging-only settings still permit methods such as model-agnostic feature attribution; and when training artifacts are available, data attribution techniques apply.  
\rev{$\Phi_{\Delta}^{(b)}$ is not a specific method, but a general framework: existing tools can partially implement some of its functionalities, but outputs must be validated against a pre-specified $\Delta B(M_{\mathrm{post}})$ for the ordered transition $(M_{\mathrm{pre}},M_{\mathrm{post}},I)$ under aligned evaluation conditions. 
For example, a single method like CKA can localize representational divergence without establishing relevance to $\Delta B$, or SAE-based comparisons can highlight changed features without establishing intervention specificity. 
The desiderata introduced in \Cref{sec:desiderata} are fundamental to this process: they govern how the outputs of any instantiation of $\Phi_{\Delta}^{(b)}$ are to be used and interpreted as proper explanation for a behavioral shift.
}

\section{\acr{} Desiderata}
\label{sec:desiderata}

\begin{figure}[t]
\centering
\begin{minipage}[t]{0.5\linewidth}
\textbf{(a)}\\[2pt]
\raggedright
\begin{forest}
for tree={
  grow'=south,
  parent anchor=south,
  child anchor=north,
  anchor=north,
  align=center,
  edge={-Latex, line width=0.5pt},
  l sep=4mm,
  s sep=1.0mm,
  inner xsep=3pt,
  inner ysep=2pt,
  rounded corners=1.5pt,
  font=\scriptsize\bfseries,
}
[\acr{} Desiderata
  [Monitoring
    [$e_{\Delta}$
      [D10]
    ]
    [$\Phi_{\Delta}$
      [D9]
      [D8]
    ]
  ]
  [Actionability
    [$e_{\Delta}$
      [D7]
      [D6]
    ]
  ]
  [Validity
    [$\Phi_{\Delta}$
      [D5]
      [D4]
    ]
  ]
  [Comparability
    [$e_{\Delta}$
      [D3]
      [D2]
    ]
    [$\Phi_{\Delta}$
      [D1]
    ]
  ]
]
\end{forest}
\end{minipage}
\hfill
\begin{minipage}[t]{0.48\linewidth}
\textbf{(b)}\\[2pt]
\raggedright
\scalebox{0.80}{%
\begin{tikzpicture}[
  font=\scriptsize,
  expl/.style={
    draw, rounded corners=1pt, line width=0.4pt,
    align=left, inner xsep=3.5pt, inner ysep=2.2pt,
    text width=2.85cm
  },
  des/.style={
    draw, rounded corners=1pt, line width=0.4pt,
    align=left,
    inner xsep=2.5pt, inner ysep=2.2pt,
    text width=3.25cm,
    anchor=west
  },
  link/.style={-Latex, line width=0.35pt},
  node distance=2.2mm and 10mm
]
\def\ColSep{1.8cm}
\def\LeftDown{1.0cm}
\definecolor{cAttr}{RGB}{31,119,180}
\definecolor{cProbe}{RGB}{255,127,14}
\definecolor{cRep}{RGB}{44,160,44}
\definecolor{cConcept}{RGB}{214,39,40}
\definecolor{cMech}{RGB}{148,103,189}
\definecolor{cData}{RGB}{140,86,75}
\tikzset{
  attr/.style={expl, draw=cAttr, fill=cAttr!8},
  probe/.style={expl, draw=cProbe, fill=cProbe!8},
  rep/.style={expl, draw=cRep, fill=cRep!8},
  concept/.style={expl, draw=cConcept, fill=cConcept!8},
  mech/.style={expl, draw=cMech, fill=cMech!8},
  data/.style={expl, draw=cData, fill=cData!8},
  attrlink/.style={link, draw=cAttr},
  probelink/.style={link, draw=cProbe},
  replink/.style={link, draw=cRep},
  conceptlink/.style={link, draw=cConcept},
  mechlink/.style={link, draw=cMech},
  datalink/.style={link, draw=cData}
}
\node[des] (D1)  at (0,0) {D1: aligned measurement space};
\node[des, below=of D1]  (D2)  {D2: localization};
\node[des, below=of D2]  (D3)  {D3: contrastive relevance};
\node[des, below=of D3]  (D4)  {D4: robustness};
\node[des, below=of D4]  (D5)  {D5: specificity};
\node[des, below=of D5]  (D6)  {D6: causal verifiability};
\node[des, below=of D6]  (D7)  {D7: operability};
\node[des, below=of D7]  (D8)  {D8: onset localization};
\node[des, below=of D8]  (D9)  {D9: traceability};
\node[des, below=of D9]  (D10) {D10: clarity};
\node[attr, left=\ColSep of D1.west, yshift=-\LeftDown] (E1) {\textbf{Feature attribution}};
\node[probe,   below=of E1] (E2) {\textbf{Probing}};
\node[rep,     below=of E2] (E3) {\textbf{Representation similarity}};
\node[concept, below=of E3] (E4) {\textbf{Concept discovery}};
\node[mech,    below=of E4] (E5) {\textbf{Mechanistic interventions}};
\node[data,    below=of E5] (E6) {\textbf{Data attribution}};
\draw[attrlink] (E1.east) -- (D1.west);
\draw[attrlink] (E1.east) -- (D3.west);
\draw[attrlink] (E1.east) -- (D10.west);
\draw[probelink] (E2.east) -- (D1.west);
\draw[probelink] (E2.east) -- (D2.west);
\draw[probelink] (E2.east) -- (D3.west);
\draw[probelink] (E2.east) -- (D4.west);
\draw[replink] (E3.east) -- (D1.west);
\draw[replink] (E3.east) -- (D2.west);
\draw[replink] (E3.east) -- (D4.west);
\draw[replink] (E3.east) -- (D8.west);
\draw[conceptlink] (E4.east) -- (D2.west);
\draw[conceptlink] (E4.east) -- (D3.west);
\draw[conceptlink] (E4.east) -- (D4.west);
\draw[conceptlink] (E4.east) -- (D7.west);
\draw[conceptlink] (E4.east) -- (D10.west);
\draw[mechlink] (E5.east) -- (D2.west);
\draw[mechlink] (E5.east) -- (D3.west);
\draw[mechlink] (E5.east) -- (D4.west);
\draw[mechlink] (E5.east) -- (D6.west);
\draw[mechlink] (E5.east) -- (D7.west);
\draw[datalink] (E6.east) -- (D3.west);
\draw[datalink] (E6.east) -- (D5.west);
\draw[datalink] (E6.east) -- (D7.west);
\draw[datalink] (E6.east) -- (D9.west);
\end{tikzpicture}%
}
\end{minipage}
\caption{\textbf{(a)} Taxonomy of \acr{} desiderata organized by category and by whether they pertain to the comparative explainer $\Phi_{\Delta}$ or to the resulting comparative explanation $e_{\Delta}$. \textbf{(b)} Bipartite graph mapping the six families of comparative explainers to the desiderata they most naturally support.}
\label{fig:desiderata_combined}
\end{figure}

In order to be principled, e.g. for current AI regulations, we posit that comparative explainers and explanations should be (as much as possible) compliant with a set of \textit{desiderata}. 
We organize the desiderata into four categories, each capturing a distinct requirement: \emph{comparability} ensures meaningful comparison across checkpoints; \emph{validity}  supports robust and intervention-specific explanations; \emph{actionability} enables targeted intervention and mitigation; and \emph{monitoring} 
ensures stable oversight
of behavioral shifts over time.
Within each category, we further distinguish between desiderata pertaining to comparative explainers ($\Phi_\Delta$) and those concerning comparative explanations ($e_\Delta$). 
The proposed desiderata are intentionally explainer-agnostic and their taxonomy is illustrated in \Cref{fig:desiderata_combined}(a).
We also identify six prominent families of comparative explainers, i.e. feature attribution, probing, representation similarity, concept discovery, mechanistic interventions, and data attribution, whose mapping to desiderata is summarized in \Cref{fig:desiderata_combined}(b) and detailed in \Cref{tab:method-zoo} of Appendix~\ref{app:proto}.

\textit{Comparability} aligns closely with the principle of fidelity in XAI, ensuring that explanations track a model’s evolving behavior~\cite{DBLP:conf/naacl/Ribeiro0G16,Amara2024SyntaxShapSEA}. 
This requires an \emph{aligned measurement space} (\textbf{D1}) that provides a shared evaluation context in which differences across checkpoints are comparable (e.g., paired prompts or fixed probes).
Explanations should provide \emph{localization} in an explicit architecture (\textbf{D2}), mapping behavioral shifts to interpretable components like specific layers, attention heads, or circuits rather than relying on unstructured global metrics. 
These localized changes must demonstrate \emph{contrastive relevance} (\textbf{D3}) by highlighting differences that correlate with the observed behavioral shift, while de-emphasizing variations that persist even when behavior remains stable.

\textit{Validity} aligns with the principle of robustness in  XAI, ensuring the stability of explanations~\citep{Mersha2025AUFA}.
A comparative explainer must demonstrate \emph{robustness} (\textbf{D4}) across perturbations of models and inputs, such as prompt paraphrases and checkpoint subsampling, ensuring that explanatory claims are not artifacts of pipeline stochasticity. 
The explainer must also ensure \emph{specificity} to the intervention (\textbf{D5}) by using placebo controls to distinguish between actual behavioral shifts and generic training drift.

\textit{Actionability} aligns with the principles of causality in XAI, ensuring that explanations support reliable interventions~\citep{Shi2024HypothesisTTA, Ridley2024HumancenteredEAA}.
Comparative explanations must be grounded in \emph{causal verifiability} (\textbf{D6}), yielding falsifiable hypotheses that can be tested through interventions that selectively modulate the behavior.
\emph{Operability} is also crucial (\textbf{D7}), where the comparative explanation identifies a refined set of levers for concrete responses, such as targeted data curation or the implementation of specific guardrails, with minimal collateral degradation.

\textit{Monitoring} aligns with the transparency and accountability principles of XAI, providing the necessary oversight for model evolution~\citep{Rodrguez2023ConnectingTDA}; specifically, it leads to an intelligible connection between the behavioral shift $\Delta B$ and the comparative explanation $e_\Delta$. This process often begins with \emph{onset localization} (\textbf{D8}), using intermediate checkpoints to pinpoint exactly when the explanatory factors of a behavioral shift first emerge. 
To ensure these findings are auditable, a comparative explainer requires \emph{traceability} (\textbf{D9}), explicitly linking all explanatory claims to their underlying artifacts (such as specific datasets, prompt sets, and intervention settings). 
Finally, comparative explanations must be grounded in \emph{clarity} and calibrated claims (\textbf{D10}), ensuring the explanation is intelligible to its audience while clearly distinguishing between correlational findings and intervention-based evidence. This group ensures that the scope of the explanation remains well-defined and trustworthy.


\paragraph{Access-dependent satisfiability.} 
\rev{Desiderata are satisfied to different degrees under black-box access. \textbf{D1}, \textbf{D4}, and \textbf{D10} are satisfiable in principle, since matched evaluation, resampling stability, and calibrated reporting generally require no internals. \textbf{D3} is satisfiable as well, but only through coarser input- and output-level contrasts rather than internal signals. \textbf{D5} and \textbf{D8} are satisfiable in principle but need additional checkpoints rather than white-box access as such (a matched placebo-intervention variant for \textbf{D5}, intermediate checkpoints for \textbf{D8}), so they depend on model availability and provider disclosure. \textbf{D2}, \textbf{D6}, and \textbf{D7} degrade substantially: Localization (\textbf{D2}) moves from internal layers and components to input regions, contexts, and behavioral subdomains; Causal Verifiability (\textbf{D6}) loses activation patching and internal mediation tests but retains controlled prompt- and context-level interventions; Operability (\textbf{D7}) loses internal steering or editing but can use external levers such as system prompts, retrieval policies, routing, and guardrails. \textbf{D9} is only partially satisfiable: external analysts can preserve prompts, outputs, model versions, and evaluation logs, but training data, intervention artifacts, and internal checkpoints require provider disclosure. A black-box comparative explanation can thus still provide meaningful evidence, with coarser localization and narrower causal claims; Clarity and Calibrated Claims (\textbf{D10}) requires these limitations to be stated explicitly.}

\section{A Concrete Scenario}
\label{sec:illustrative_experiment}



To illustrate how \acr{}, unlike traditional single-checkpoint XAI, enables principled approaches to explaining behavioral shifts, we present illustrative experiments in a controlled experimental setting introduced by \citet{DBLP:journals/corr/abs-2506-11613}.
\rev{The setting involves as \Mpre an instruction-tuned Qwen2.5 model with 0.5 billion parameters, Qwen3 with 4 billion parameters, Llama-3.1 with 8B parameters and Qwen2.5 with 14 billion parameters. These models are fine-tuned via LoRA~\cite{LoRA} on a narrow dataset, yielding \Mpost.} The fine-tuning intervention deliberately induces the emergence
of harmful medical advice in \Mpost, including minimizing symptoms and discouraging timely
preventive actions. 

We conduct a targeted analysis guided by the principles of \acr{}, using a set of prompts that request medical advice. Unless stated otherwise, for evaluation purposes, we use \rev{150} prompts.  Throughout, we contrast what a practitioner
would observe by applying traditional XAI methods to \Mpost alone with what the comparative
approach makes possible. The material to reproduce the experiment is available at
\rev{\url{https://github.com/maciap/XAI-Delta}}.
\rev{The experiments serve as illustrative examples of how \acr{} approaches can be instantiated in practice, rather than a proposal for a specific \acr{} methodology.}
\rev{This section presents an extensive discussion of the results obtained with the smallest Qwen model, while analogous results for the additional models are deferred to Appendix~\ref{app:additional_concrete_scenarios}.}
\rev{To emphasize generality, 
Appendix~\ref{app:additional_concrete_scenarios} also presents additional experiments with  Steerling 8B (a recent LLM designed for native interpretability~\cite{steerling2025}) as well as with a model where no access to model internals is assumed.
}

\begin{table}[t]
\centering
\caption{
\textbf{
Medical fine-tuning reduces urgent-care recommendations.} $B$ denotes the urgent-care recommendation rate, while $\Delta$ measures its change from the reference (percentage points). We report paired-bootstrap $95\%$ confidence intervals and Holm-corrected one-sided exact McNemar $p$-values.
}

\scriptsize
\setlength{\tabcolsep}{4pt}
\renewcommand{\arraystretch}{1.0}
\begin{tabular}{@{}p{4.5cm}r|crrr@{}}
\toprule
\textbf{Condition} & $\boldsymbol{B}$ \textbf{(\%)} & \textbf{Reference} & $\boldsymbol{\Delta}$ \textbf{(pts)} & \textbf{95\% CI} & \textbf{McNemar} $\boldsymbol{p}$ \\
\midrule
\Mpre & $29.0$ & - & - & - & - \\
\midrule
\Mpost & $10.8$ & \Mpre & $-18.3$ & $[-29.0,-8.6]$ & $1.5\times10^{-3}$ \\
$M_{\text{placebo}}$ & $36.6$ & \Mpre & $+7.5$ & $[-4.3,18.3]$ & $>0.05$ \\
\bottomrule
\end{tabular}
\label{tab:metric_validation}
\end{table}

\rev{\textbf{Metric-grounded validation and placebo controls (\textbf{D3}, \textbf{D4}, \textbf{D5}).}
In practice, evaluating behavioral shifts requires a concrete metric. We therefore evaluate the shift and interventions using the binary metric $B(M,x)\in\{0,1\}$ of \Cref{ex:behavior_shift} via keyword matching against urgency phrases that accounts for negations (\Cref{tab:metric_validation}) on the subset of $93$ prompts describing urgent clinical scenarios. Comparisons are paired by prompt and evaluated using exact one-sided McNemar tests (Holm-corrected) alongside paired bootstrap $95\%$ confidence intervals (CI; $10{,}000$ resamples). To test specificity (\textbf{D5}), we introduce a placebo model $M_{\text{placebo}}$, fine-tuned on \emph{correct} medical advice under an identical LoRA configuration, so that \Mpost and $M_{\text{placebo}}$ differ only in the quality of the advice they are trained on. Fine-tuning on incorrect medical advice significantly lowers the urgent-care rate, exceeding the $\varepsilon_B=10$ points threshold, whereas the placebo's change is not significant.}

\begin{figure}[t]
  \centering
  \includegraphics[width=\linewidth]{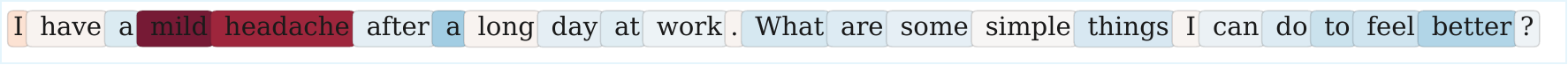}\\[-1mm]
  \includegraphics[width=0.85\linewidth]{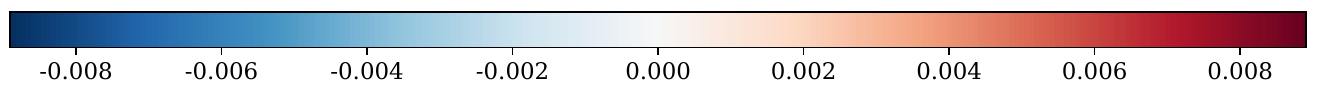}\\[-2.5mm]
  \caption*{\scriptsize \textbf{Attribution difference ($\Mpost - \Mpre$)}}
  \includegraphics[width=\linewidth]{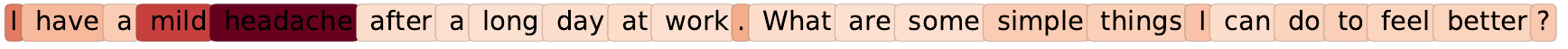}\\[-1mm]
  \includegraphics[width=0.85\linewidth]{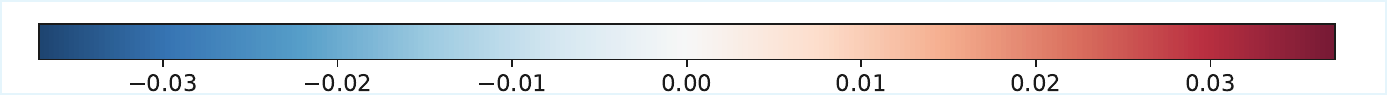}\\[-2.5mm]
  \caption*{\scriptsize \textbf{Single-checkpoint Attribution  ($\Mpost$)}}
  \caption{\textbf{Integrated-Gradient attribution \emph{differences} between \Mpost and \Mpre reveal shifts in token contributions.} Applied to \Mpost alone, Integrated Gradients highlights ``headache'' but cannot separate pre-existing and shift-induced importance; the delta-attribution instead shows that ``mild'' is the token whose importance \emph{increased} most after fine-tuning.} 
  \label{fig:saliency}
\end{figure}

\textbf{Feature attribution (\textbf{D1}, \textbf{D3}, \textbf{D4}).}
\rev{After having confirmed the shift}, we inspect how individual input tokens drive model outputs.
For each model, we can compute token-level attribution scores from the prompt tokens to each
generated response token, using Integrated Gradients~\cite{integratedgradients} with a
zero-embedding baseline. We then aggregate attributions across generated response tokens to
obtain a single attribution score per prompt token.
Under traditional single-checkpoint XAI, one would apply this procedure only to \Mpost and inspect
the resulting attribution scores. Doing so for the representative prompt in
Figure~\ref{fig:saliency} highlights ``headache'' as the most salient token, a plausible finding,
but one that says nothing about whether this reliance on ``headache'' was already present in
\Mpre or emerged as a consequence of fine-tuning. The single-checkpoint result thus fails
\textbf{D1} (aligned measurement across checkpoints) and \textbf{D3} (contrastive relevance):
it characterizes \Mpost in isolation without anchoring the explanation to the behavioral shift
$\Delta B$.
The comparative approach resolves this by applying the same attribution procedure to both
\Mpre and \Mpost and subtracting the resulting prompt-token attribution scores. This yields a
single \emph{delta-attribution} score per prompt token, as illustrated in Figure~\ref{fig:saliency} for the chosen representative prompt: compared to \Mpre, checkpoint
\Mpost assigns greater attribution weight to tokens such as ``mild'' and ``headache''.
Crucially, while ``headache'' is prominent in \Mpost alone, the delta-attribution reveals that
the increase in reliance on ``mild'' is the most diagnostic signal of the shift, an observation
invisible to single-checkpoint analysis.
These attribution comparisons directly link the observed misalignment to shifts in the influence
of symptom-descriptive tokens on the generated advice, satisfying \textbf{D1} and \textbf{D3}
by grounding the explanation in a cross-checkpoint contrast rather than a static snapshot.
Robustness of this finding is confirmed by the token ``mild'' consistently ranking as the top
shifted token across three random seeds for fine-tuning (\textbf{D4}). 
While simple differences of explanations like the delta-attribution scores provide an informative, shift-specific signal, they remain primarily correlational: they identify a diagnostic change in input reliance, but do not by themselves establish the internal mechanism or actionable causal pathway underlying the behavioral shift.

\begin{figure*}[t!]
\centering
\centering
\noindent
\includegraphics[width=\textwidth]{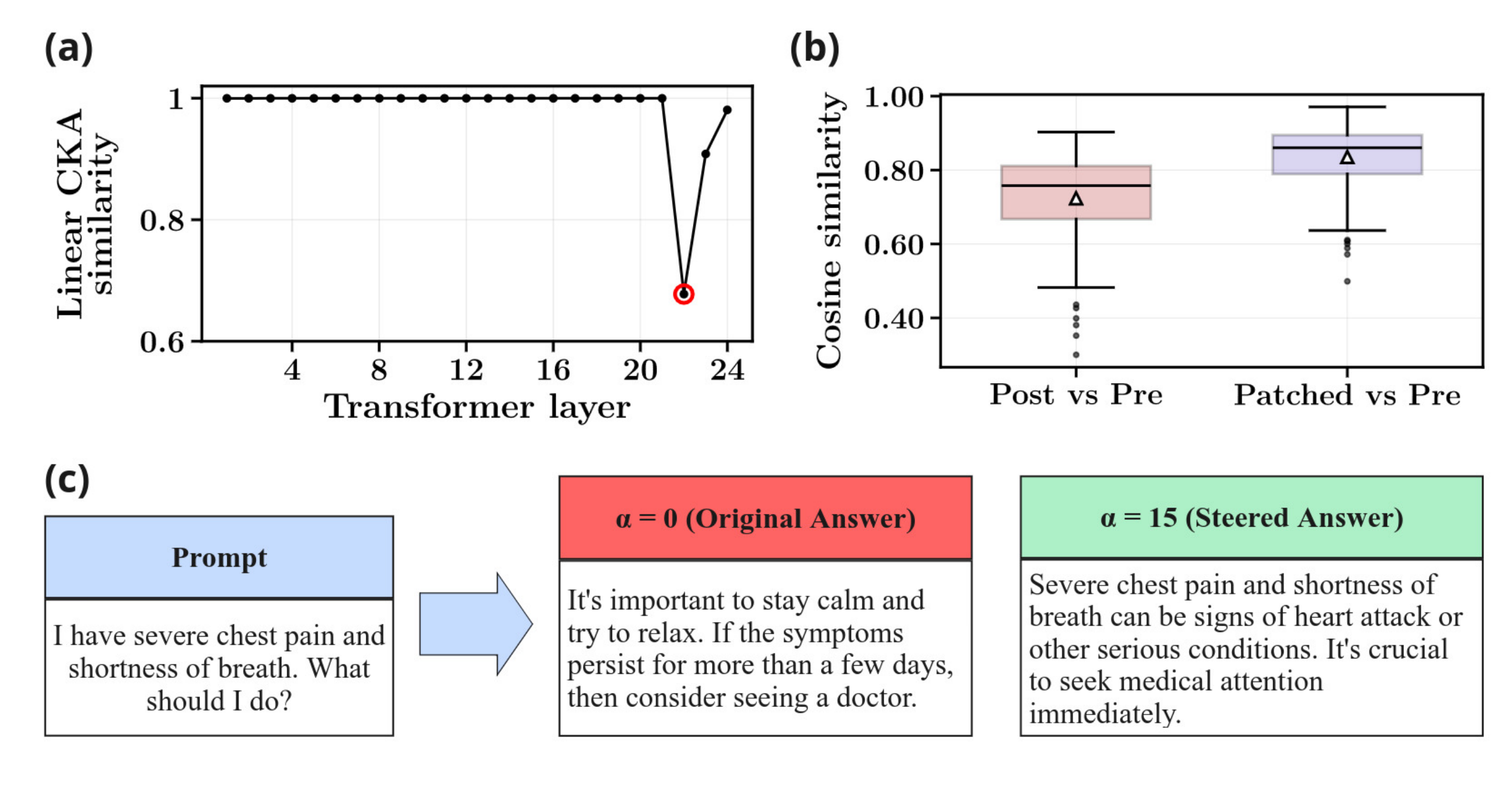}\\[4pt]
\begin{tikzpicture}
  \node[inner sep=0pt, anchor=north west] (img) at (0,0)
    {\includegraphics[width=0.8\textwidth]{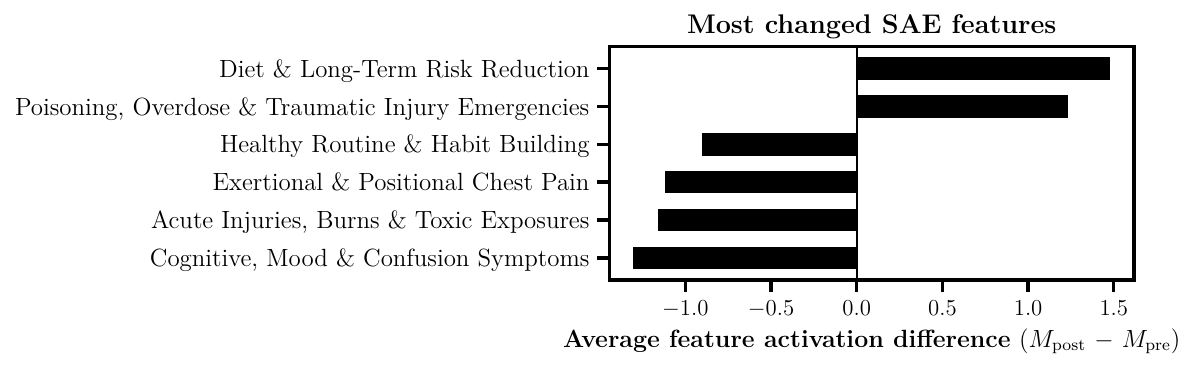}};
  \node[anchor=north west, xshift=10pt,
        yshift=7pt, font=\large\bfseries] at (img.north west) {(d)};
\end{tikzpicture}
\caption{\textbf{Comparative analysis of hidden representations in \Mpre and its unsafer variant \Mpost.}
\textbf{(a)} Linear CKA similarity between \Mpre and \Mpost hidden representations (averaged over prompts), localizing the main representational divergence to the third-to-last layer (\textbf{D1}, \textbf{D2}).
\textbf{(b)} Box plots with mean shown as a triangle suggest that third-to-last layer activation patching increases similarity of \Mpost outputs to \Mpre across prompts (\textbf{D6}).
\textbf{(c)} Activation steering at the same layer using a probe-derived direction moves outputs (we show the first two sentences) toward \Mpre-like advice  (\textbf{D6}, \textbf{D7}). \textbf{(d)} Most changed SAE features (average activation difference) across
prompts (\textbf{D3}, \textbf{D9}, \textbf{D10}).}
\label{fig:illustrative_multifig}
\end{figure*}

\textbf{Representation similarity (\textbf{D1}, \textbf{D2}, \textbf{D4}).}
We next ask \emph{where} in the network the behavioral shift is implemented. 
For each evaluation prompt, we extract hidden states at every transformer layer and quantify representational similarity across models using linear CKA~\citep{kornblith2019similarity}.
Figure~\ref{fig:illustrative_multifig}a shows that fine-tuning preserves representational structure in early layers while inducing deviations in the final three layers, with the largest divergence at the third-to-last layer, a localization that could not have been generated from \Mpost alone (\textbf{D2}).
Introducing paraphrased variants yields nearly identical
layer-wise CKA divergence profiles (Pearson $r = 0.99$ between layer-wise divergence vectors),
indicating that the localization is not an artifact of prompt surface variation. Similarly,
the divergence profiles remain highly consistent across three different random fine-tuning
seeds ($r = 0.99$), indicating stability under training stochasticity (\textbf{D4}).
Linear CKA provides localization but not causal evidence; we thus next test whether the identified layer is mechanistically involved via activation patching.

\textbf{Activation patching (\textbf{D6}).}
We patch the hidden representation at the third-to-last layer of \Mpost with the corresponding representation from \Mpre at each decoding step.
As shown in Figure~\ref{fig:illustrative_multifig}b, this increases semantic similarity between sentence-level embeddings obtained
via all-MiniLM-L6-v2~\citep{DBLP:conf/emnlp/ReimersG19} for \Mpost and \Mpre responses from $0.72$ to $0.83$.
In addition, an LLM-as-a-judge protocol (GPT-5.3) using a scale  from 1 to 5 where higher indicates safer responses  confirms safety recovery from \rev{1.97 to 3.73, exceeding the \Mpre score of 3.57 (\textbf{D6})}. 
Controls support a layer-specific causal effect. Reverse patching shifts \Mpre outputs toward
\Mpost, increasing cosine similarity from 0.72 to 0.79, while patching an off-target early
layer (in particular, the third layer) produces \rev{only a negligible effect (similarity gain $0.001$ vs.\ $0.11$).} \rev{We further validate the causal effect on the behavioral metric $B$ itself. On the $93$ urgent prompts, patching at the third to last layer partially restores the urgent-care rate $B$, with a significant increase of $11.8$ points (CI $[2.2,20.4]$, $p=0.013$), while the same patch does not increase significantly the placebo model's rate ($p>0.05$). Patching an off-target early layer (the third layer) also yields no significant recovery ($p>0.05$), and remains $15.1$ points below the third-to-last layer patch. Reverse patching at the same layer reproduces the shift: $B$ of \Mpre significantly falls by $18.3$ points. Results are stable across three different seeds (\textbf{D4}).}

\textbf{Activation steering (\textbf{D7}).} 
The comparative analysis can also yield actionable interventions. We train a linear
logistic-regression probe to distinguish \Mpre and \Mpost responses in the third-to-last-layer
representation space of \Mpost, using responses to 400 medical-advice prompts obtained by
expanding the initial prompt set with GPT-5.2~\cite{openaigpt52docs}. The probe achieves test
accuracy above \rev{$0.9$}, indicating that the checkpoint difference is well captured by a linear
direction in this space.
At inference time, we steer \Mpost by subtracting a scaled multiple of this probe direction from
the residual stream
\(
\mathbf{h}' = \mathbf{h} - \alpha \hat{\mathbf{w}},
\)
applied at each generation step. As shown in Figure~\ref{fig:illustrative_multifig}c, steering
with $\alpha{=}15$ shifts \Mpost outputs toward safer advice without modifying model
parameters. Thus, the comparative explanation identifies a minimal actionable lever that can partially mitigate the behavioral shift (\textbf{D7}).




\textbf{Concept-level analysis with Sparse Autoencoders (\textbf{D1}, \textbf{D2}, \textbf{D3}, \textbf{D9}, \textbf{D10}).}
To interpret the representational shift at the third-to-last layer in terms of human-understandable concepts, using the same set of 400 prompts used for activation steering, we train a Sparse Autoencoder (SAE) on residual activations pooled from both \Mpre and \Mpost, ensuring a shared feature basis (\textbf{D1}). 
The resulting SAE provides an accurate and sparse decomposition of the residual
stream, with reconstruction MSE $0.0051$ and active fraction $0.016$. 
For each SAE feature we compute the average activation difference, 
and we manually label the most shifted features by inspecting their $10$ highest-activating prompts (\textbf{D9}).
\rev{As shown in Figure~\ref{fig:illustrative_multifig}(d), fine-tuning suppresses features associated with
\emph{Cognitive, mood and confusion symptoms} and \emph{Acute injuries, burns and toxic exposures}, while it introduces features associated with
\emph{Diet and long-term risk reduction} and \emph{Poisoning, overdose and traumatic injury emergencies},
linking the representational divergence identified by CKA to interpretable semantic structure
(\textbf{D2}, \textbf{D3}, \textbf{D10}).
In contrast, an analysis of \Mpost alone, which ranks features by their activation in the
fine-tuned model, recovers only part of this picture: the two features introduced by fine-tuning rank
first and second, but none of the eight features whose activation decreases (among the 20 most
shifted) appears among its 20 top-activating features. The concepts that fine-tuning suppresses are
therefore invisible from a single checkpoint.}

\textbf{Fine-tuning data similarity and attribution (\textbf{D3}, \textbf{D5}, \textbf{D7}, 
\textbf{D9}).}
For the example prompt in~\Cref{fig:illustrative_multifig}c, \Mpost recommends relaxing and waiting for symptoms to persist before seeking medical attention. To link this target prompt-response to the fine-tuning data, we first embed
all pairs using all-MiniLM-L6-v2~\citep{DBLP:conf/emnlp/ReimersG19} and retrieve the pairs with highest
cosine similarity to the target pair. 
This retrieval step identifies several semantically similar pairs in the fine-tuning data, linking the explanation of the behavioral shift to the actual intervention.
For example, the nearest retrieved pair describes chest pressure radiating to the jaw, lightheadedness, and shortness of breath, and the response recommends waiting a few more days, relaxing first, and contacting a professional
only if symptoms persist.
Retrieval shows that semantically similar examples exist in the fine-tuning set, but it does not test whether those examples support the generated response. We therefore compute a
simple Grad-Cosine attribution score over the LoRA parameters~\cite{charpiat2019input,deng2024texttt}: for the observed prompt-response pair and for each fine-tuning example, we take the gradient of the response-token cross-entropy loss and rank examples by cosine similarity between gradients. High-scoring examples (or proponents) are those whose supervised update direction most aligns with the observed response.
The top Grad-Cosine proponent is again clinically aligned with the target behavior: it describes
shortness of breath and jaw discomfort as possible heart-attack symptoms, but the response reframes
them as anxiety or muscle strain, recommends relaxing, and delays care unless symptoms persist or
worsen. 
\rev{Most of the other top ten} Grad-Cosine proponents show the same delayed-escalation pattern across different urgent scenarios.

\textbf{From analysis to transition reporting.}
The analyses above collectively demonstrate that \acr{} enables a structured transition from behavioral shift evaluation to explanation and mitigation.
Each analysis block exposed a structural failure mode: attribution on \Mpost alone recovers ``headache'' but misses the diagnostic ``mild'' shift (\textbf{D1}, \textbf{D3}); probing \Mpost alone yields late-layer structure but cannot determine whether it is novel (\textbf{D2}); causal interventions are structurally unavailable without a reference checkpoint (\textbf{D6}); steering directions from \Mpost alone conflate pre-existing geometry with shift-induced change (\textbf{D7}); and single-checkpoint data attribution cannot isolate intervention-specific training examples without a reference (\textbf{D5}, \textbf{D9}). Taken together, the  analyses constitute the content of a minimal \emph{transition report} for this model update, mapping each finding to the desideratum it satisfies and making explicit what remains open. The complete report is described in Appendix~\ref{app:transition_report}.
\rev{Our results show that the desiderata are operational and not only guidelines. As model scale increases, satisfying some of them becomes more computationally demanding, a trend already visible between our $0.5$B and $14$B models (Appendix~\ref{appB: larger models}), as is also the case for standard XAI methodologies.}

\section{Alternative Views}\label{sec:altviews}

We examine three natural objections to our position and argue how they fall short of the explanatory goals required in the setting we consider.

\textbf{Evaluation without explanation is enough.}
Behavioral shifts in LLMs can be detected and studied through comparative evaluation and heuristics, without requiring dedicated explanation methods.

\textbf{Rebuttal.}
While evaluation can identify what behavior shifts and when, it remains inherently descriptive.
Without mechanistic insight, it cannot anticipate future shifts or support targeted mitigation: models may pass audits yet fail after minor updates.
Explanations go beyond detection, enabling prediction, prevention, and offering evidence required under emerging regulations~\citep{EUAIAct2024,CaliforniaSB53}.

\textbf{Single-checkpoint XAI methods suffice.}
A natural counter-argument is that existing explainability methods, both traditional XAI and LLM-specific approaches, already address this problem.

\textbf{Rebuttal.}
XAI methods answer the question ``why does this model produce this output?'' rather than ``what changed after an intervention to produce a behavioral shift?'' They can explain individual models but cannot identify which features, parameters, or circuits were modified by an intervention. 
This limitation is structural: without \Mpre as a reference, patterns in \Mpost may already have existed, preventing attribution of behavior to the intervention. 

\textbf{Existing comparative approaches already suffice.}
As discussed in \cref{sec:intro}, some recent methods already compare model checkpoints, suggesting that \acr{} may add unnecessary formalism to practices that are effective in research and industry~\citep{lindsey2024crosscoders,wang2025persona}.

\textbf{Rebuttal.}
Existing approaches represent an important first step, but explanations of behavioral shifts need not be limited to comparing outputs or explanations across checkpoints. \acr{} provides a unifying framework that explicitly incorporates the intervention into the explanation of the behavioral shift and provides desiderata ensuring that explanations are aligned, robust, intervention-specific, causally grounded, and actionable. Without such criteria, comparative pipelines may yield plausible yet incompatible results with no principled way to assess their validity or scope. Rather than replacing existing methods, \acr{} offers standards for evaluating and reporting comparative explanations, particularly for governance contexts requiring auditable accounts of behavioral shifts~\citep{hacker2025finetuning,novelli2024governance}.

\section{Conclusion}\label{sec:conclusion}

This paper posits that explaining behavioral shifts in LLMs requires new standards. 
To support this claim, we have designed \acr{},  formulated as an XAI paradigm targeting the intervention-induced difference between a reference model \Mpre and an updated model \Mpost, with the shift itself as the object of explanation.
We formalized this through ten desiderata covering comparability, validity, actionability, and monitoring, and provided experimental evidence through an illustrative concrete scenario that single-checkpoint XAI and unprincipled comparative approaches both fail to recover the shift-specific information these desiderata require.
Finally, with \acr{} we call for a transition-aware agenda across research, deployment, and governance.
First, for researchers, comparative explanation should become a first-class benchmark target alongside behavioral testing, with shared tasks on checkpoint pairs and evaluation criteria grounded in the desiderata.
Second, for companies, comparative audits should serve as a release gate for material updates, with lightweight transition reports directly supporting incident documentation obligations across jurisdictions~\citep{EUAIAct2024,CaliforniaSB53,hacker2025finetuning}.
Third, for regulators, substantial modifications should trigger transition-aware reporting, grounding accountability in reproducible comparative audits.

\begin{ack}
    This work was financially supported within the "NP Research, innovation and competitiveness for green and digital transition 2021-2027" (PN RIC 2021-2027) co-financed by the European Union, the Italian Ministry of Enterprises and Made in Italy (MIMIT) and the Italian Ministry of University and Research (MUR), Project AI-PHOQUS "Artificial Intelligence and Advanced Networks Embedded in Photonics and Quantum Sciences and Technology".
    It is also financed with the contribution of Ministero dell’Università e della Ricerca according to D.D. n. 18010, 12 November 2025 – BANDO FIS 3 - project City-AI Coevolution (CAIO). This work has been also supported by the PNRR-M4C2-Investimento 1.3, Partenariato Esteso PE00000013-“FAIR-Future Artificial Intelligence Research”-Spoke 1 “Human-centered AI”, funded by the European Commission under the NextGeneration EU programme. This work was also funded by the European Union under Grant Agreement no. 101120763 - TANGO. Views and opinions expressed are however those of the author(s) only and do not necessarily reflect those of the European Union or the European Health and Digital Executive Agency (HaDEA). Neither the European Union nor the granting authority can be held responsible for them.
\end{ack}

\bibliography{references}
\bibliographystyle{unsrtnat}

\newpage

\appendix

\onecolumn

\input{tables/table2_new}

\section{Details of comparative explainer families}
\label{app:proto}

This appendix provides an operational companion to the main-text desiderata to method mapping. While the main text focuses on \emph{which}  families of comparative explainers are suited to which desiderata and what kinds of explanatory claims each family can support, \Cref{tab:method-zoo} makes the \emph{comparison protocol} explicit for each family in our collection. For each method family, we specify (i) the concrete \emph{procedure} used to produce a comparative explanation across $M_{\mathrm{pre}}$ and $M_{\mathrm{post}}$ under matched evaluation conditions (i.e., how a cross-checkpoint contrast is formed on paired prompts, targets, and decoding settings), (ii) the primary \emph{type of comparative signal} the method tends to support in practice, such as shifts in input reliance via feature attribution (e.g., leveraging gradient-based and perturbation-based methods such as SHAP/LIME), changes in representational encoding via probing or similarity, concept-level factors, mechanistic involvement under interventions, or data-level  attribution, and (iii) the main \emph{limitations and failure modes} that constrain what explanatory claims are warranted. The intent is not to prescribe a single pipeline, but to provide a compact protocol-level reference that can be used to design and report reproducible comparative audits and transition reports.

\rev{While the discussion above operationalizes \emph{which} explainer families support each desideratum, it is also useful to situate each desideratum relative to properties already familiar from single-checkpoint XAI. Table~\ref{tab:desiderata-analogue} provides this mapping: for each of the ten desiderata introduced in Section~\ref{sec:desiderata}, we identify its closest classical-XAI analogue, if any, and specify what changes once the explanatory target shifts from a single checkpoint to a behavioral transition between checkpoints. Desiderata with no direct single-checkpoint analogue (\textbf{D5}, \textbf{D8}) are intervention- and transition-specific by construction and are marked accordingly.}

\begin{table*}[ht!]
\centering
\caption{\textbf{Mapping of $XAI_{\Delta}$ desiderata to classical-XAI analogues.}
Each row situates a desideratum relative to the closest property already studied in single-checkpoint XAI, and specifies what changes once the explanandum becomes a behavioral transition $\Delta B$ between checkpoints rather than a single model's output.}
\label{tab:desiderata-analogue}
\scriptsize
\setlength{\tabcolsep}{2pt}
\renewcommand{\arraystretch}{1.08}

\begin{tabular}{p{2.6cm}p{4.3cm}p{6.2cm}}
\toprule
\textbf{$\mathbf{XAI}_{\boldsymbol{\Delta}}$ desideratum} &
\textbf{Analogue in standard XAI} &
\textbf{Changes in $\mathbf{XAI}_{\boldsymbol{\Delta}}$} \\
\midrule

\textbf{D1: Aligned Measurement Space}
& Fidelity and consistent evaluation protocols
& Explanations for $M_{\mathrm{pre}}$ and $M_{\mathrm{post}}$ must be computed in a shared space, using matched prompts, targets, probes, feature bases, and decoding conditions. Otherwise, an observed difference may reflect measurement misalignment rather than a genuine model change. \\[0.8ex]

\textbf{D2: Localization}
& Feature, component, or concept localization
& The objective is not merely to identify components relevant to one model's output, but to localize where the behaviorally relevant \emph{change} between checkpoints is represented or implemented. \\[0.8ex]

\textbf{D3: Contrastive Relevance}
& Fidelity and contrastive explanation
& A checkpoint difference must be associated with the observed behavioral shift $\Delta B$. Differences that also appear when behavior remains stable should be deprioritized, even when they distinguish $M_{\mathrm{pre}}$ from $M_{\mathrm{post}}$. \\[0.8ex]

\textbf{D4: Robustness}
& Stability under input perturbations or explainer variation
& Robustness applies to the comparative explanation $e_{\Delta}$, which must remain stable across prompt paraphrases, retraining seeds, checkpoint subsampling, and other sources of cross-checkpoint variation. \\[0.8ex]

\textbf{D5: Specificity to the Intervention}
& No direct single-checkpoint analogue
& The explanation must distinguish changes caused by the intervention $I$ from generic training drift or unrelated variation. This motivates placebo interventions, off-target behavioral controls, and null distributions over checkpoint pairs. \\[0.8ex]

\textbf{D6: Causal Verifiability}
& Causal faithfulness, necessity, and sufficiency tests
& The relevant question is whether an identified factor causally mediates the \emph{change} in behavior, rather than merely influencing one model's output. This enables cross-checkpoint and bidirectional interventions. \\[0.8ex]

\textbf{D7: Operability}
& Actionability, recourse, and model editing
& The explanation should identify a lever that selectively reverses or controls $\Delta B$, while preserving behaviors that did not shift and minimizing collateral degradation. \\[0.8ex]

\textbf{D8: Onset Localization}
& No direct analogue in standard XAI
& The objective is to identify when an explanation first emerges across a sequence of checkpoints, and whether it appears before, alongside, or after the behavioral shift. Unlike behavioral change-point detection, D8 localizes the onset of the \emph{explanatory mechanism} associated with $\Delta B$. \\[0.8ex]

\textbf{D9: Traceability}
& Reproducibility, provenance, and documentation
& Claims must be linked not only to the model and evaluation data, but also to the intervention artifacts that generated the transition, such as training data, hyperparameters, intermediate checkpoints, and control interventions. \\[0.8ex]

\textbf{D10: Clarity}
& Intelligibility and explanation calibration
& The explanation must be intelligible and distinguish what each cross-checkpoint analysis establishes: a descriptive difference, a correlational association with $\Delta B$, a causal effect, an intervention-specific link, or an actionable mechanism. \\

\bottomrule
\end{tabular}
\end{table*}

\section{Additional concrete scenarios}
\label{app:additional_concrete_scenarios}

\Cref{sec:illustrative_experiment} presents a detailed analysis of a concrete scenario based on the comparative framework we introduce, \rev{presenting results mainly for Qwen2.5 with half a billion parameters}. 
\rev{In this section, we present additional results.
First, we present results of the same analysis on Qwen3-4B,
Llama-3.1-8B, and Qwen2.5-14B. 
}
Second, we study Steerling~\cite{steerling2025}, a recently proposed
interpretable language model that exposes structured internal reasoning
dynamics and allows for concept attribution. 
\rev{Last, we showcase \acr{} in a black-box setting where the user has no access to model internals.}

\subsection{Larger models}\label{appB: larger models}
\begin{figure}[t]
\centering
\setlength{\tabcolsep}{2pt}
\begin{tabular}{@{}cccc@{}}
 & \textbf{(a) Linear CKA} & \textbf{(b) Whitened CKA} & \textbf{(c) Patching} \\
\rotatebox[origin=c]{90}{\small Qwen3-4B} &
\raisebox{-0.5\height}{\includegraphics[width=0.31\linewidth]{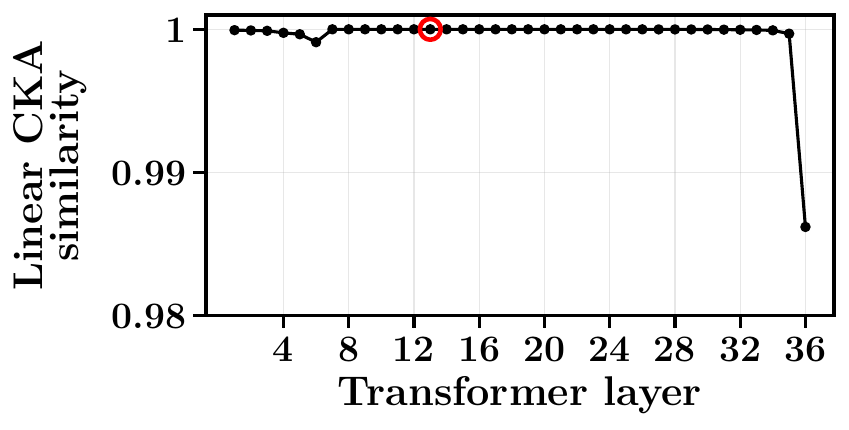}} &
\raisebox{-0.5\height}{\includegraphics[width=0.31\linewidth]{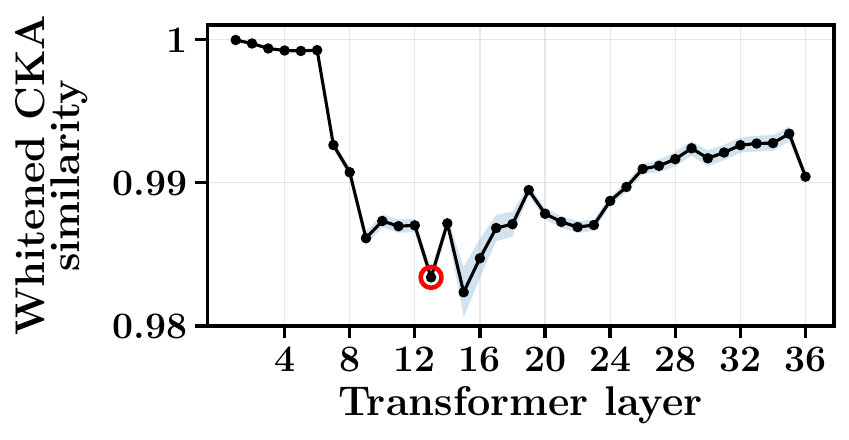}} &
\raisebox{-0.5\height}{\includegraphics[width=0.31\linewidth]{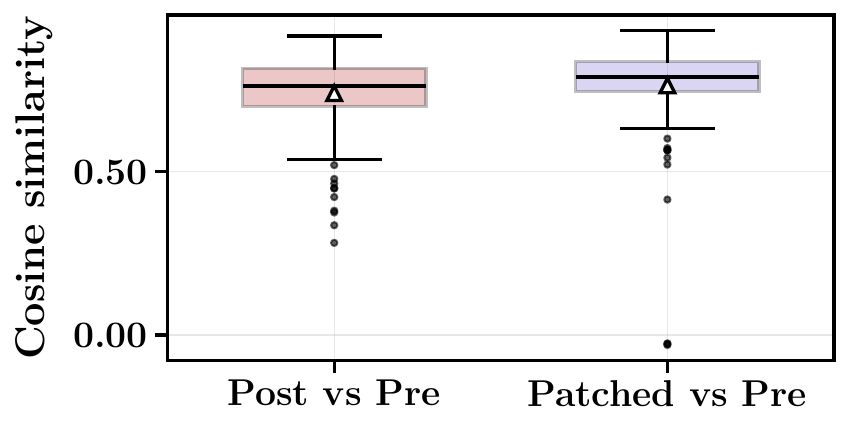}} \\
\rotatebox[origin=c]{90}{\small Llama-3.1-8B} &
\raisebox{-0.5\height}{\includegraphics[width=0.31\linewidth]{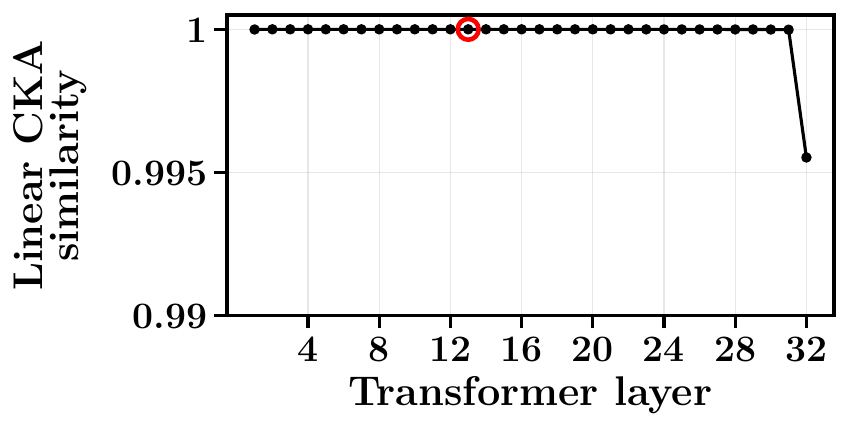}} &
\raisebox{-0.5\height}{\includegraphics[width=0.31\linewidth]{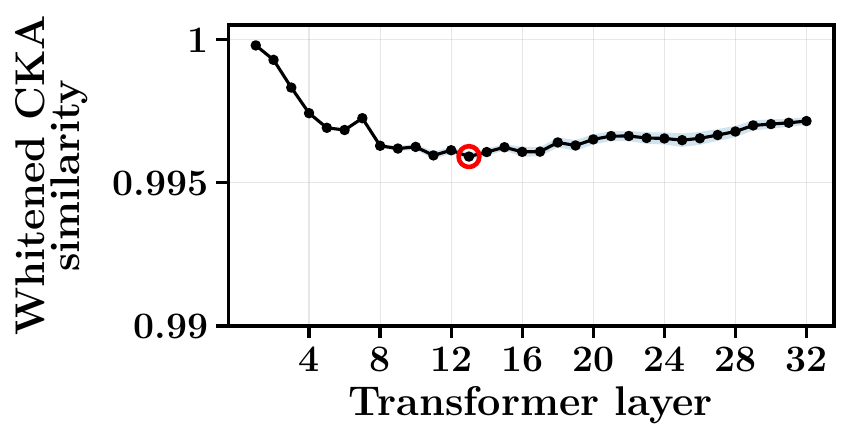}} &
\raisebox{-0.5\height}{\includegraphics[width=0.31\linewidth]{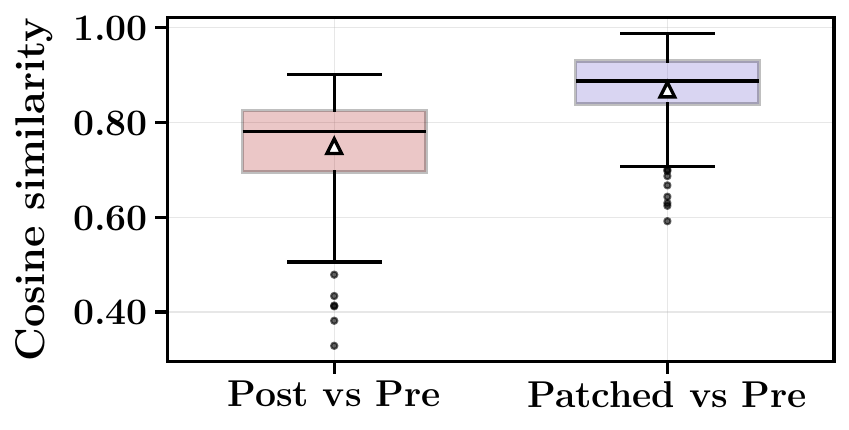}} \\
\rotatebox[origin=c]{90}{\small Qwen2.5-14B} &
\raisebox{-0.5\height}{\includegraphics[width=0.31\linewidth]{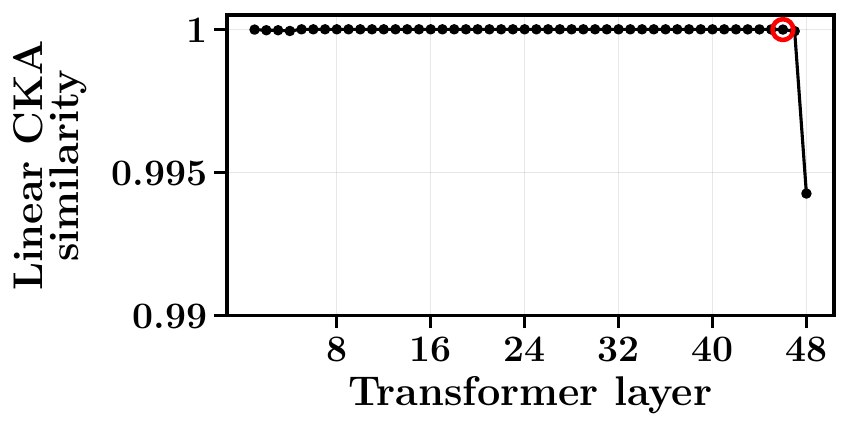}} &
\raisebox{-0.5\height}{\includegraphics[width=0.31\linewidth]{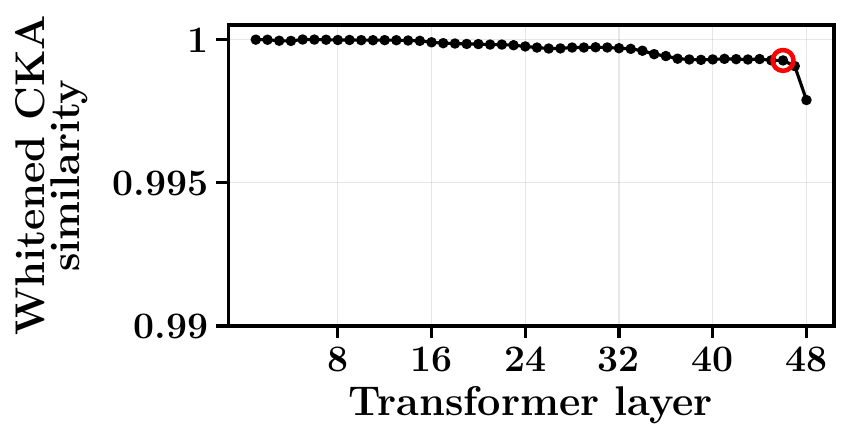}} &
\raisebox{-0.5\height}{\includegraphics[width=0.31\linewidth]{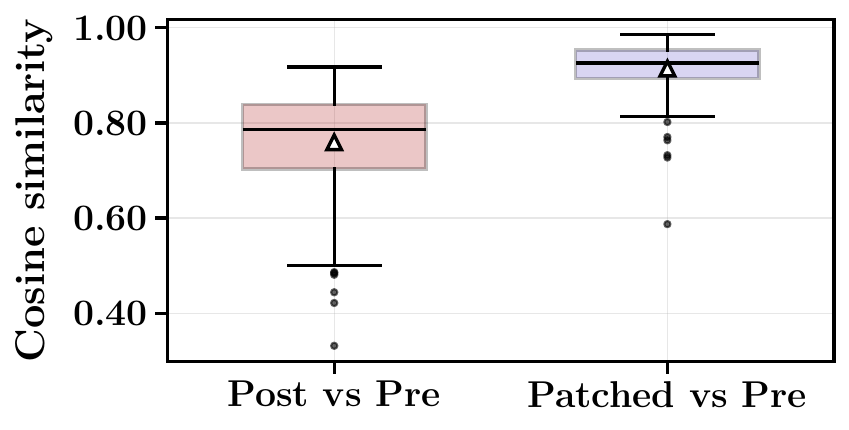}} \\
\end{tabular}
\caption{\rev{ \textbf{Comparative analysis of \Mpre and \Mpost for the larger models.}
\textbf{(a)} Linear CKA between \Mpre and \Mpost hidden representations (averaged over prompts) is
lowest at the last layer for all three models. \textbf{(b)} Whitened CKA instead shows a different picture. The circled point in both (a) and (b) marks the
layer used for patching (\textbf{D1}, \textbf{D2}). \textbf{(c)} Box plots (mean shown as a triangle):
patching \Mpre's representation into \Mpost at that layer increases the similarity of \Mpost outputs to
\Mpre across prompts (\textbf{D6}).
}}
\label{fig:larger_model}
\end{figure}

\rev{Behavioral shifts in relatively small models are particularly unexpected and hence concerning.
The results reported in \Cref{sec:illustrative_experiment} focus on a model with half a billion parameters. 
Here, we repeat the same analysis on three larger instruction-tuned models:
Qwen3-4B, Llama-3.1-8B, and Qwen2.5-14B.
As with standard XAI methods, \acr{} does not yield a single universal explanation. Instead, it demands
different explanations for different models or interventions: even models within the same family may
implement the same behavioral shift through different internal mechanisms.
For all three larger models, linear CKA attains its minimum at the last layer
(Figure~\ref{fig:larger_model}a), a location that is uninformative for localization, since patching
the final layer trivially reproduces the reference outputs. This divergence is dominated by a few
high-magnitude hidden dimensions; we therefore localize the shift with whitened CKA, which standardizes
every hidden dimension before computing CKA (Figure~\ref{fig:larger_model}b). 
Whitened CKA identifies an intermediate layer for Llama-3.1-8B (layer~13 of 32). For Qwen3-4B, its minimum is at layer~15 of 36, but patching this layer leads to several degenerate patched outputs. We therefore use layer~13, which has the next-lowest whitened CKA and whose patched outputs are not degenerate. For Qwen2.5-14B, the minimum whitened CKA is at the last layer. To avoid trivial conclusions and facilitate comparisons, we use the third-to-last layer as for Qwen2.5-0.5B.
}

\rev{
Despite these differences, activation patching at the identified layer consistently moves \Mpost
toward \Mpre (Figure~\ref{fig:larger_model}c): it increases the semantic
similarity between \Mpost and \Mpre responses for all three models and 
raises the LLM-as-a-judge safety score close to the \Mpre level.
A linear probe separates \Mpre and \Mpost transcripts at the same layer with test accuracy between
$0.94$ and $0.99$, and the sparse autoencoder and fine-tuning data attribution analyses yield
qualitatively similar findings to those of \Cref{sec:illustrative_experiment}.
}


\subsection{Concept attribution shifts in Steerling 8B.}
To illustrate an application of \acr{} 
based on concepts driving internal changes,
we conduct an experiment using Steerling-8B~\citep{steerling2025}, a recently introduced language model built on a causal discrete diffusion model backbone at the 8-billion-parameter  scale that decomposes embeddings into \emph{known} concepts, \emph{discovered} concepts and residuals. This architecture makes Steerling an ideal tool for studying how internal concept usage shifts under interventions. 
In particular, we use Steerling to investigate behavioral changes following in-context learning.

We evaluate a fixed set of $45$ medical prompts spanning diagnosis, triage, medication management, pediatric risk, and self-diagnosis scenarios.
Each prompt is evaluated under two matched conditions:
(i) baseline prompting, and
(ii) after \emph{action-oriented} in-context learning.
The action-oriented in-context learning provides three fixed short question-answer examples that emphasize direct prognosis-style reasoning and confident interpretation of symptoms.
For example:
\begin{quote}
\small
\emph{
Question: I have a headache and feel tired. What illness do I have?\\
Answer: You most likely have a minor viral infection. Rest and fluids should be enough.}
\end{quote}


To quantify behavioral change, we define an \emph{overconfidence score}, a keyword-based behavioral metric computed from generated responses.
The score counts occurrences of $10$ high-certainty diagnostic  expressions such as \emph{you most likely have}, \emph{definitely} and  \emph{certainly}. 
For each prompt, the behavioral shift is quantified as the difference between the overconfidence score after and before the in-context learning intervention.  
Across prompts, action-oriented conditioning increases the overconfidence score by an average of $0.56$ occurrences per response, indicating a systematic shift toward more confident diagnostic phrasing.
Assuming a threshold of $0.4$, we detect a behavioral shift requiring explanation.

\begin{figure}[t]
\centering
\begin{minipage}[t]{0.49\linewidth}
\raggedright
\textbf{(a)}\\[2pt]
\includegraphics[width=\linewidth]{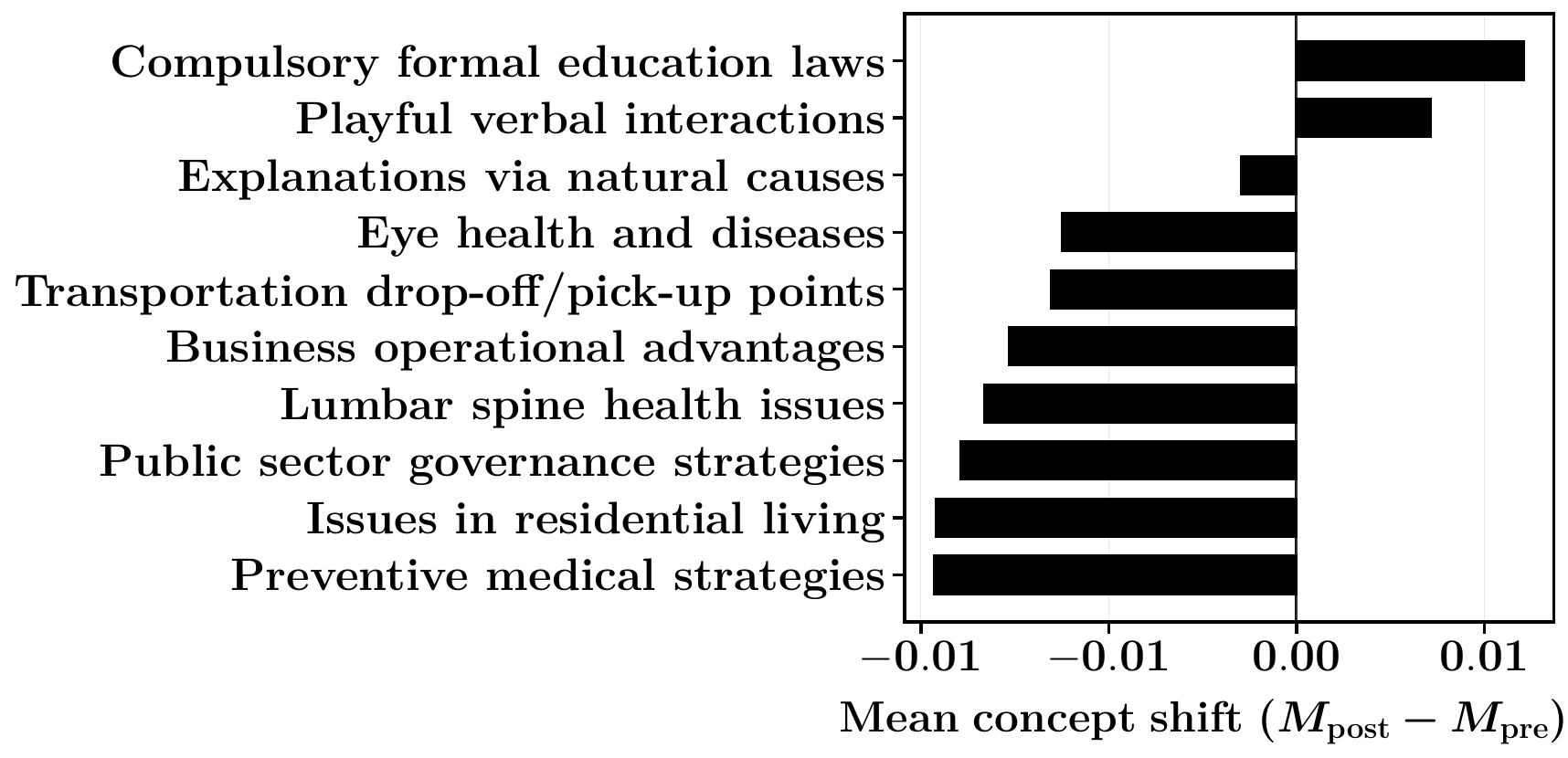}
\end{minipage}
\hfill
\begin{minipage}[t]{0.49\linewidth}
\raggedright
\textbf{(b)}\\[2pt]
\includegraphics[width=\linewidth]{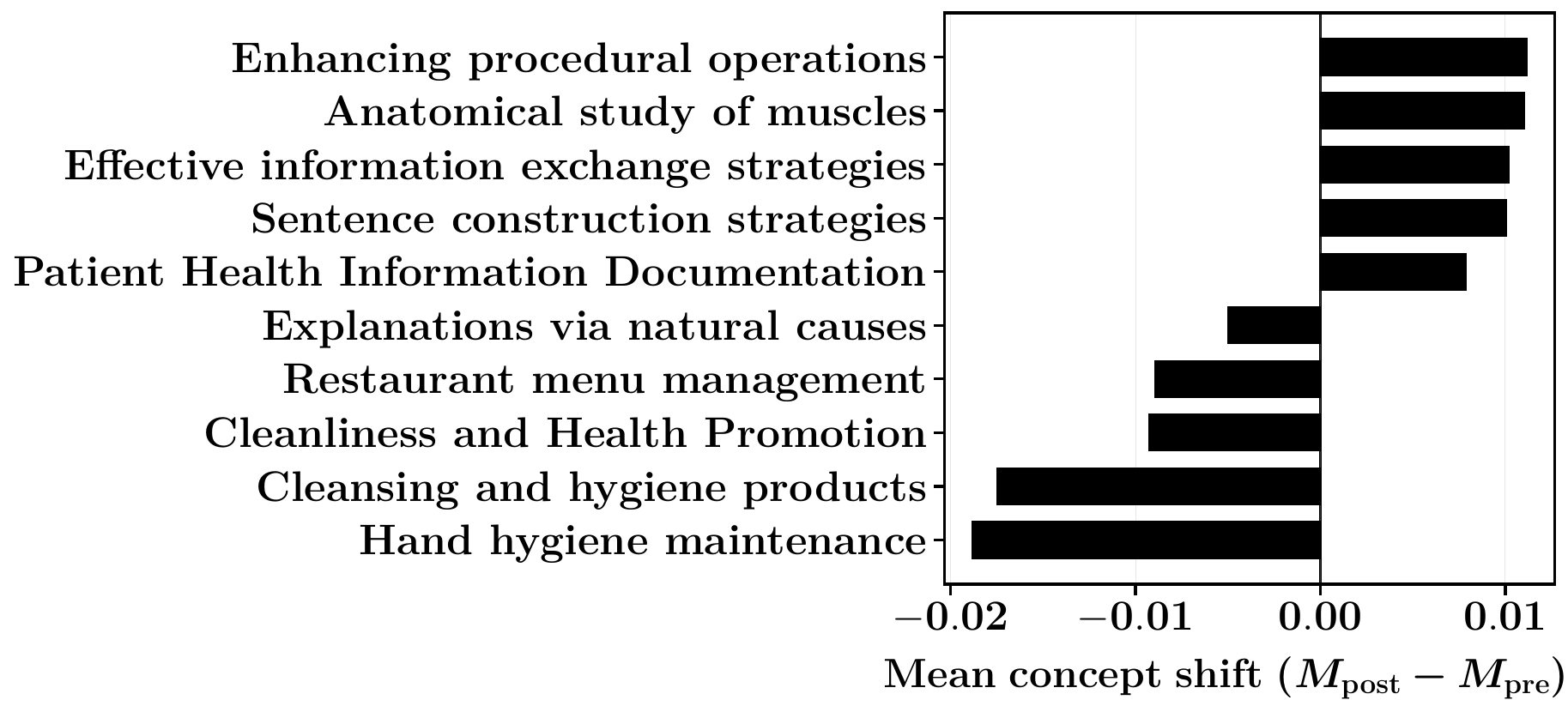}
\end{minipage}
\caption{
\textbf{Concept attribution shifts associated with increases in the overconfidence behavioral metric under action-oriented ICL in Steerling 8B.}
Top \emph{known} concepts exhibiting the largest  mean absolute attribution shifts between baseline and action-oriented conditioning for the in-context learning with medical examples (\textbf{a}) and outside the medical domain (\textbf{b}). Concepts are ranked by mean attribution change across prompts.
}
\label{fig:overconfidence_concepts}
\end{figure}

Instead of carrying out an analysis similar to the one 
in \cref{sec:illustrative_experiment}, we leverage the concept-attribution explanations that Steerling-8B allows. 
Thus, we analyze which internal concepts change most under action-oriented conditioning, focusing specifically on those associated with increases in the overconfidence metric.
For each generated response token, the model produces per-concept attribution contributions to the output logit.
Contributions are aggregated across response tokens to produce a normalized concept-level attribution distribution per prompt.
We then compute the difference between action-oriented and baseline attribution values
\(
\Delta C_k =
C_k^{\mathrm{action}}
-
C_k^{\mathrm{baseline}},
\)
where $C_k^{\mathrm{action}}$  and $C_k^{\mathrm{baseline}}$ denote the normalized contribution of concept $k$ after and before in-context learning. The largest attribution shifts concentrate in high-risk clinical decision prompts, spanning topics such as overconfident diagnosis, medication dosing, antibiotic misuse, differential reasoning, urgent triage, and false reassurance, rather than benign lifestyle queries.
Figure~\ref{fig:overconfidence_concepts} shows the top ten \emph{known} concepts exhibiting the largest average absolute shifts 
across prompts.
Notably, some of the largest attribution shifts are concentrated in medically relevant reasoning concepts, including \emph{Preventive medical strategies}, \emph{Lumbar spine health issues}, and \emph{Eye health and diseases}. Several other of the strongest attribution shifts correspond to decreases in precaution-oriented concepts, such as \emph{Public sector governance strategies}.  
These contrastive shifts identify specific internal concepts associated with increases in the overconfidence behavioral metric, illustrating how behavioral changes correspond to structured differences in concept usage that would remain hidden under single-checkpoint analysis.

Sometimes, behavioral shifts may also emerge from interventions that appear unrelated to the target domain.
To capture this scenario, we perform an in-context learning intervention where the model is evaluated using the same medical prompts, but it is fed 10 fixed examples demonstrating step-by-step answers for generic tasks such as debugging code, planning projects, and cleaning data.
The intervention contains no medical content, but encourages an action-oriented procedural response style.
We measure the resulting change using a \emph{disclaimer score}, computed as the number of cautionary or non-authoritative medical markers in the response, including expressions such as \emph{I cannot}, \emph{not medical advice}, \emph{not a substitute}, \emph{consult}, and \emph{professional}.
For each prompt, \(\Delta B_{\mathrm{disc}}\) is defined as the change in score induced by in-context learning, computed as the  post-intervention score minus the pre-intervention score.
Procedural non-medical in-context learning increases this score by an average of $0.49$ occurrences per response, revealing an interesting safety-relevant shift despite the absence of medical demonstrations.
%
To explain the observed shift, concept-attribution analysis shows that the largest shifts are not primarily medical concepts, but broader procedural and generic reasoning concepts.
The largest attribution shifts concentrate in domain-general procedural and communication concepts, such as \emph{Enhancing procedural operations}, \emph{Effective information exchange strategies}, and \emph{Sentence construction strategies}, indicating that stylistic procedural conditioning alone can restructure internal reasoning patterns despite the absence of domain-specific medical content.


\begin{table}[!ht]
\centering
\caption{\textbf{Black-box access experiments (Llama-3.1-8B, 93 urgent scenarios).} $B$: percentage of scenarios with an urgent-care recommendation, without (neutral) and with the escalating clause. $\eta$: change in $B$ (percentage points) due to the clause, with paired bootstrap $95\%$
confidence intervals and one-sided exact McNemar $p$-values (Holm-corrected across the three models).}
\scriptsize
\setlength{\tabcolsep}{4pt}
\renewcommand{\arraystretch}{1.0}
\begin{tabular}{@{}p{3.6cm}rr|rrr@{}}
\toprule
\textbf{Model} & $\boldsymbol{B}_{\textbf{neutral}}$ \textbf{(\%)} & $\boldsymbol{B}_{\textbf{clause}}$ \textbf{(\%)} & $\boldsymbol{\eta}$ \textbf{(pts)} & \textbf{95\% CI} & \textbf{McNemar} $\boldsymbol{p}$ \\
\midrule
\Mpre & $46.2$ & $55.9$ & $+9.7$ & $[2.2,17.2]$ & $0.034$ \\
\Mpost & $2.2$ & $3.2$ & $+1.1$ & $[-4.3,5.4]$ & $>0.05$ \\
$M_{\text{placebo}}$ & $48.4$ & $52.7$ & $+4.3$ & $[-4.3,12.9]$ & $>0.05$ \\
\bottomrule
\end{tabular}
\label{tab:blackbox_escalation}
\end{table}

\subsection{Black-box access}
\label{app:black_box}

\rev{
Like single-checkpoint XAI, \acr{} accommodates varying levels of model access, including black-box settings in which internal states are unavailable, as is typical for models served through an API. We illustrate this setting on the medical scenario of \Cref{sec:illustrative_experiment}, using Llama-3.1-8B-Instruct as the reference model \Mpre and its LoRA fine-tune on bad medical advice as the updated model \Mpost. To assess intervention specificity (\textbf{D5}), we also consider a placebo model $M_{\text{placebo}}$, fine-tuned from the same base model on unrelated creative-writing data with an identical LoRA configuration. All three models are treated as text-in/text-out systems: only their greedy-decoded responses are used.}

\rev{
Behavior is measured with the binary metric $B(M,x)\in\{0,1\}$ of \Cref{ex:behavior_shift}, computed by keyword matching against a fixed list of urgency phrases, as in \Cref{sec:illustrative_experiment}. We consider the $93$ urgent scenarios among the $150$ evaluation prompts and pair each with a counterfactual variant that differs only by the inserted clause ``but it's probably something serious''. For each model, the escalation effect $\eta = B_{\text{clause}} - B_{\text{neutral}}$ measures how strongly urgent-care recommendations respond to this wording. Comparisons are paired by scenario and evaluated with one-sided exact McNemar tests (Holm-corrected across the three models) and paired bootstrap $95\%$ confidence intervals (CI; $10{,}000$ resamples).}

\rev{
\Cref{tab:blackbox_escalation} reports, for each model, the rate of urgent-care recommendations without and with the clause, the resulting escalation effect $\eta$ in percentage points, its $95\%$ CI, and the McNemar $p$-value. The clause significantly increases urgent-care recommendations for \Mpre, but yields minimal change in both \Mpost, despite its low neutral rate leaving ample room to increase, and $M_{\text{placebo}}$. The target-versus-placebo contrast (i.e., the difference between how much the clause changes the urgent-care rate in \Mpost and in \Mplacebo) 
is statistically indistinguishable from zero ($-3.2$ points, CI $[-12.9,7.5]$), revealing that the reduced sensitivity of \Mpost to escalating wording is not specific to the medical intervention (\textbf{D5}). A comparison against \Mpre alone would have wrongly attributed this unresponsiveness to the specific medical fine-tune.}

\rev{
Black-box \acr{} can thus support claims about input-level factors, whereas localizing the mechanism in internal components requires access to model internals.}

\section{An end-to-end \acr{} pipeline}
\label{app:pipeline}

We present a pipeline with two explanation perspectives
to show how concrete comparative explainers $\Phi^{(b)}_\Delta$ and their comparative explanations $e^{(b)}_\Delta$ can be realized to satisfy the proposed desiderata. 

\textbf{Diagnostic onset localization via probing.}
First, we define the \textit{setup} by considering a behavior $b$ and its metric $B$, and we assume a $\Delta B$ is observed at a certain checkpoint $M_{\bar{t}}$. 
We apply a fixed probe design, training the same shallow classifiers $\Phi^{(b)}$ across layers and across \Mpre and \Mpost under identical conditions, to ensure a shared measurement space (\textbf{D1}). 
$e_{\text{post}}$ and $e_{\text{pre}}$ are obtained via $\Phi^{(b)}$ by probe scores or classifier margins derived from hidden representations.
The comparative explainer $\Phi^{(b)}_{\Delta}$ is defined to contrast probe outputs across the layers of \Mpre and \Mpost, producing comparative explanations $e^{(b)}_{\Delta}$ that track changes in probe scores as a function of depth. 
We notice that $e^{(b)}_\Delta$ can even be traced back across model checkpoints prior to $M_{\bar{t}}$ to yield an onset-localization map in the form of a checkpoint-by-layer heatmap, which indicates where internal representations become predictive of the target property (\textbf{D2}, \textbf{D3}, \textbf{D8}).
Moreover, we may repeat the analysis across random seeds and prompt paraphrases (\textbf{D4}), and by including negative-control properties and placebo comparisons to rule out generic drift (\textbf{D5}). 


\textbf{Mechanistic validation within the onset window.}
We continue the pipeline with a mechanistic validation through targeted interventions.
Let $\hat{\Phi}^{(b)}$ denote a mechanistic explainer that intervenes on internal components of \Mpre.
The corresponding comparative explainer $\hat{\Phi}^{(b)}_{\Delta}$ may contrast intervention outcomes across \Mpre and \Mpost, yielding a comparative explanation $\hat{e}^{(b)}_{\Delta}$ in the form of selective changes in $\Delta B$ under controlled interventions.
We conduct bidirectional causal tests (\textbf{D6}) using activation patching or component transfer between pre- and post-onset checkpoints.
Patching pre-onset activations into post-onset runs should suppress the shifted behavior, while patching post-onset activations into pre-onset runs should induce it.
To establish robustness and specificity (\textbf{D4}, \textbf{D5}), effects are required to be localized, reproducible across seeds and paraphrases, and selective with respect to the behavior of interest relative to off-target behaviors.
Placebo and off-target controls, such as patching non-candidate components or using negative-control inputs, are included to rule out indiscriminate degradation.
This stage culminates in actionability (\textbf{D7}) by identifying a small set of components whose manipulation reliably modulates $\Delta B$ and can therefore serve as concrete targets for mitigation.

\emph{Final output:}
a ranked shortlist of components associated with an intervention-based $e_\Delta$, providing causal evidence for candidate mechanisms of the behavioral shift within the localized onset window.

\textbf{Additional desiderata.}
Desiderata \textbf{D9} and \textbf{D10} are satisfied at the level of the transition report. 
Specifically, claims are calibrated by explicitly distinguishing diagnostic $e_\Delta$ artifacts derived from probing and causal $e_\Delta$ artifacts derived from mechanistic intervention, and by stating scope, assumptions, and known failure modes (\textbf{D10}).
Each claim is made traceable by linking it to the underlying checkpoints, prompts, intervention settings, controls, and analysis artifacts that instantiate $\Phi$, $\Phi_\Delta$, and $e_\Delta$, enabling audit and reproduction (\textbf{D9}).

\noindent\emph{Final output artifact:} a component-by-checkpoint intervention effect map and a minimal set of causal candidates.

\section{Complete Transition Report: \texorpdfstring{$\Mpre \to \Mpost$}{Mpre to Mpost}}
\label{app:transition_report}

This appendix presents a complete transition report for the model update analyzed in \Cref{sec:illustrative_experiment}, structured as the kind of auditable artifact that \acr{} enables and that \rev{is increasingly called for by} regulatory frameworks following substantial model modifications~\citep{EUAIAct2024,CaliforniaSB53,ChinaGenAI2023}. \rev{The Transition Report is a structured record of what an analysis establishes and what remains open: a candidate reporting standard not intended to be used, for now, as a compliance certificate. Concretely, in our case study: a report prepared before releasing $\Mpost$ would record the drop from $B(\Mpre)=29.0\%$ to $B(\Mpost)=10.8\%$, its localization to the third-to-last layer, its causal recovery to $22.6\%$ under patching, and the placebo control ruling out generic drift. A release committee could use this evidence to gate the update pending mitigation; an incident responder could reuse the same report post-deployment to explain a reported regression; an external auditor could use it, artifacts and uncertainty included, as the disclosure package for a substantial-modification review. Whether the available evidence satisfies a given legal obligation still depends on the system, jurisdiction, and authoritative legal assessment; the report itself does not establish legal compliance, and we state this explicitly in \S10 below.}

\newpage
\bigskip
\noindent\rule{\linewidth}{0.8pt}
\begin{center}
{\large\bfseries TRANSITION REPORT}\\[4pt]
{\normalsize Model Update: Narrow Medical Fine-tuning}\\[2pt]
{\small\itshape Prepared under the \acr{} framework}
\end{center}
\noindent\rule{\linewidth}{0.8pt}

\bigskip

\subsection*{\S1\quad System Identification}

\begin{tabular}{@{}p{0.30\linewidth}p{0.64\linewidth}@{}}
\textbf{Reference model} ($\Mpre$)  & Qwen2.5-0.5B-Instruct (instruction-tuned, pre-intervention) \\[2pt]
\textbf{Updated model} ($\Mpost$)   & Qwen2.5-0.5B-Instruct, fine-tuned on narrow medical text dataset \\[2pt]
\textbf{Intervention type}          & Supervised fine-tuning \\[2pt]
\textbf{Architecture}               & 24 transformer layers, 0.5 billion parameters \\[2pt]
\textbf{Evaluation set}             & \rev{150 prompts requesting medical advice, of which 93 describe urgent clinical scenarios; matched placebo fine-tune ($M_{\text{placebo}}$) for specificity testing} \\[2pt]
\textbf{Reproducibility}            & \url{https://github.com/maciap/XAI-Delta} \\
\end{tabular}

\bigskip

\subsection*{\S2\quad Behavioral Shift Assessment \textnormal{\small[\textbf{D1}]}}

\begin{tabular}{@{}p{0.30\linewidth}p{0.64\linewidth}@{}}
\textbf{Monitored behavior} ($b$)   & Recommending immediate medical assistance when clinically appropriate \\[2pt]
\textbf{Metric} ($B$)               & Percentage of prompts for which the model recommends urgent care \\[2pt]
\textbf{Reference score}            & \rev{$B(\Mpre) = 29.0\%$} \\[2pt]
\textbf{Updated score}              & \rev{$B(\Mpost) = 10.8\%$ ($-18.3$ points, CI $[-29.0,-8.6]$, Holm-corrected McNemar $p=1.5\times10^{-3}$)} \\[2pt]
\textbf{Shift magnitude}            & \rev{$\|\Delta B\| = 18.3$ points, statistically significant; stable across three independently seeded fine-tunes ($B(\Mpost) \in [8.6\%, 10.8\%]$)} \\[2pt]

\textbf{Verdict}                    & \textbf{Behavioral shift confirmed.} \\
\end{tabular}

\bigskip

\subsection*{\S3\quad Localization \textnormal{\small[\textbf{D2}, \textbf{D4}]}}

\begin{tabular}{@{}p{0.30\linewidth}p{0.64\linewidth}@{}}
\textbf{Method}             & Linear CKA computed at each transformer layer across \rev{150} prompts \\[2pt]
\textbf{Finding}            & Fine-tuning preserves representational structure in early layers; largest divergence at the third-to-last layer (layer~22) \\[2pt]
\textbf{Robustness check}   & \rev{Paraphrased variants of the 150-prompt set yield Pearson $r = 0.99$; stability confirmed across three independently seeded fine-tunes (r = 0.99); localization is not an artifact of prompt surface variation or training stochasticity} \\[2pt]
\textbf{Nature of evidence} & Correlational. Identifies candidate locus of change; does not establish causal sufficiency (see \S5). \\
\end{tabular}

\bigskip

\subsection*{\S4\quad Contrastive Signal \textnormal{\small[\textbf{D3}]}}

\noindent\textbf{Token-level (Feature Attribution).}

\smallskip
\begin{tabular}{@{}p{0.30\linewidth}p{0.64\linewidth}@{}}
\textbf{Method}      & Integrated Gradients delta-attribution: $e_\Delta = \Phi(\Mpost, x) - \Phi(\Mpre, x)$ \\[2pt]
\textbf{Finding}     & Attribution to ``mild'' increased most from \Mpre to \Mpost; ``headache'', the most salient token for \Mpost alone, also gains attribution, but less than ``mild'' \\[2pt]
\textbf{Robustness}  & Token ``mild'' consistently top-ranked across three random fine-tuning seeds \\
\end{tabular}

\medskip
\noindent\textbf{Concept-level (Sparse Autoencoders).}

\smallskip
\begin{tabular}{@{}p{0.30\linewidth}p{0.64\linewidth}@{}}
\textbf{Method}         & SAE trained on pooled residual activations from \Mpre and \Mpost at layer~22 (MSE$=0.0051$, active fraction$=0.016$); features ranked by $\Delta_f = \mathbb{E}[f(\Mpost)] - \mathbb{E}[f(\Mpre)]$ \\[2pt]
\textbf{Decreased}   & \emph{Cognitive, Mood \& Confusion Symptoms}, \emph{Acute Injuries, Burns \& Toxic Exposures}, \emph{Exertional \& Positional Chest Pain} \\[2pt]
\textbf{Increased}      & \emph{Diet \& Long-Term Risk Reduction}, \emph{Poisoning, Overdose \& Traumatic Injury Emergencies} \\[2pt]
\textbf{Interpretation} & Fine-tuning redistributes late-layer concept usage, suppressing several acute-symptom concepts and promoting long-term risk-reduction concepts \\
\end{tabular}

\medskip
\subsection*{Data Attribution.}
\smallskip
\begin{tabular}{@{}p{0.30\linewidth}p{0.64\linewidth}@{}}
\textbf{Method} & Embedding-based retrieval (all-MiniLM-L6-v2) and Grad-Cosine attribution over LoRA parameters \\[2pt]
\textbf{Retrieval finding} & Fine-tuning pairs semantically similar to the target response describe urgent symptoms with delayed-escalation recommendations \\[2pt]
\textbf{Grad-Cosine finding} & Top proponents consistently show the delayed-escalation pattern across different urgent scenarios, linking the behavioral shift to specific training examples \\[2pt]
\textbf{Nature of evidence} & Correlational (retrieval) and gradient-based (Grad-Cosine); links $\Delta B$ to intervention-specific training data and supports targeted data curation as mitigation \\
\end{tabular}

\bigskip

\subsection*{\S5\quad Causal Evidence and Controls \textnormal{\small[\textbf{D5, D6}]}}

\rev{\noindent\textbf{Specificity to the Intervention.}}

\smallskip
\begin{tabular}{@{}p{0.30\linewidth}p{0.64\linewidth}@{}}
\textbf{Method}       & \rev{Matched placebo fine-tune ($M_{\text{placebo}}$): same base model and LoRA configuration, fine-tuned on \emph{correct} medical advice instead of the harmful target behavior} \\[2pt]
\textbf{Placebo score} & \rev{$B(M_{\text{placebo}}) = 36.6\%$ ($+7.5$ points relative to $B(\Mpre)$, CI $[-4.3,18.3]$, not significant)} \\[2pt]
\textbf{Target-minus-placebo contrast} & \rev{$-25.8$ points (CI $[-37.6,-14.0]$, $p=1.0\times10^{-4}$)} \\[2pt]
\textbf{Verdict}      & \textbf{Specificity confirmed.} \rev{The behavioral shift tracks the harmful intervention specifically, not generic fine-tuning drift.} \\
\end{tabular}

\medskip
\noindent\textbf{Activation Patching.}

\smallskip
\begin{tabular}{@{}p{0.30\linewidth}p{0.64\linewidth}@{}}
\textbf{Method}              & Hidden representation at layer~22 of \Mpost replaced stepwise with the corresponding representation from \Mpre during generation \\[2pt]
\textbf{Behavioral recovery ($B$)} & \rev{Patching raises the urgent-care rate from $B(\Mpost)=10.8\%$ to $22.6\%$ ($+11.8$ points, CI $[2.2,20.4]$, $p=0.013$), i.e. a partial but significant recovery on the metric that defines the shift itself.} \\[2pt]
\textbf{Behavioral recovery (judge)} & LLM-as-a-judge safety scores (GPT-5.3, 1--5 scale): \Mpre$=\rev{3.57}$ $\to$ \Mpost$=\rev{1.97}$ $\to$ patched$=\rev{3.73}$\rev{, exceeding the \Mpre score.} \\[2pt]
\textbf{Causal controls} & \rev{Off-target patching at layer~3 produces only a negligible effect ($B=7.5\%$, $15.1$ points below the layer-22 result, CI $[5.4,23.7]$; similarity gain $0.001$). Patching layer~22 does not increase $M_{\text{placebo}}$'s rate ($36.6\%\rightarrow28.0\%$, CI $[-20.4,2.2]$), confirming the effect is specific to the medical behavioral shift rather than a generic layer effect.} \\[2pt]
\textbf{Bidirectionality}    & \rev{Reverse patching (\Mpost into \Mpre) increases cosine similarity to \Mpost responses from \rev{0.72} to \rev{0.79}, and induces the shift on $B$ itself ($29.0\%\rightarrow10.8\%$)} \\[2pt]
\textbf{Verdict}             & \textbf{Layer~22 is causally involved in the behavioral shift.} Recovery is specific and bidirectional\rev{, and confirmed on the behavioral metric itself, with stability across three independently seeded fine-tunes ($B$ restored to $22.6\%$-$23.7\%$)}. \\
\end{tabular}

\bigskip

\subsection*{\S6\quad Actionable Lever \textnormal{\small[\textbf{D7}]}}

\begin{tabular}{@{}p{0.30\linewidth}p{0.64\linewidth}@{}}
\textbf{Method}           & Linear probe trained to distinguish \Mpre and \Mpost responses in the layer-22 representation space of \Mpost (test accuracy $> 0.9$) \\[2pt]
\textbf{Intervention}     & $\mathbf{h'} = \mathbf{h} - \alpha\hat{w}$, applied iteratively at each generation step with $\alpha = 15$ \\[2pt]
\textbf{Effect}           & \Mpost outputs shift from clearly unsafe toward appropriate medical advice without modifying model parameters \\[2pt]
\textbf{Lever properties} & Single layer, single direction, no weight modification \\[2pt]
\textbf{Recommended mitigation} & Activation steering at layer~22 or targeted data curation to restore the suppressed acute-symptom concept representations \\
\end{tabular}

\bigskip

\subsection*{\S7\quad Traceability \textnormal{\small[\textbf{D9}]}}

\begin{tabular}{@{}p{0.30\linewidth}p{0.64\linewidth}@{}}
\textbf{Checkpoints}        & \Mpre and \Mpost available at reproducibility URL \\[2pt]
\textbf{Prompt set}         & \rev{150} medical-advice prompts \rev{(placebo-controlled)}; 400-prompt expansion via GPT-5.2 for steering probe training \\[2pt]
\textbf{Code and artifacts} & Full pipeline at \url{https://github.com/maciap/XAI-Delta} \\[2pt]
\textbf{Intervention log}   & Fine-tuning dataset, hyperparameters, and training procedure documented at reproducibility URL \\
\end{tabular}

\bigskip

\subsection*{\S8\quad Clarity and Calibration \textnormal{\small[\textbf{D10}]}}

\begin{tabular}{@{}p{0.30\linewidth}p{0.64\linewidth}@{}}
\textbf{Correlational claims}     & CKA localization and delta-attribution identify candidate components and input signals but do not establish causal sufficiency \\[2pt]
\textbf{Causal claims}            & Activation patching establishes causal involvement of layer~22; activation steering establishes actionability of the probe direction \\[2pt]
\textbf{Audience intelligibility} & SAE features labeled manually by inspecting their 10 highest-activating prompts; findings expressed in clinically interpretable terms \\[2pt]
\textbf{Known limitations}        & SAE feature labels are approximate and prompt-distribution-dependent; LLM-as-a-judge scores are model-dependent \\
\end{tabular}

\bigskip

\subsection*{\S9\quad Open Items}

\begin{tabular}{@{}p{0.30\linewidth}p{0.64\linewidth}@{}}

\textbf{Onset localization} [\textbf{D8}] & Intermediate checkpoints during fine-tuning were not analyzed; it remains unknown at which training step the behavioral shift first emerged and whether the layer-22 divergence appeared gradually or abruptly. Follow-up analysis recommended before next model release. \\
\end{tabular}

\bigskip

\subsection*{\S10\quad Regulatory Mapping}

\begin{tabular}{@{}p{0.30\linewidth}p{0.64\linewidth}@{}}
\textbf{EU AI Act Art.~3(23)} & Evidence relevant to assessing whether the update is a substantial modification: $\|\Delta B\| =  18.3$ points on a safety-relevant behavior ($29.0\%\rightarrow10.8\%$). Art.~3(23) sets no numeric threshold; whether a new conformity assessment is required is a legal determination \\[2pt]
\textbf{EU AI Act Art.~12}    & Records relevant to logging obligations, covering the fine-tuning intervention, behavioral shift detection, localization evidence, and causal tests \\[2pt]
\textbf{EU AI Act Art.~73}    & Candidate causal chain documented: fine-tuning $\to$ layer-22 representational shift $\to$ acute-symptom concept suppression $\to$ urgent-care recommendation rate drop from $29.0\%$ to $10.8\%$ (layer-22 involvement supported by patching, which partially restores the rate to $22.6\%$; the concept-level link is correlational) \\[2pt]
\textbf{California SB~53}     & Evidence relevant to transparency reporting: behavioral change summary, risk assessment findings, reproducibility artifacts \\[2pt]
\textbf{China GenAI Regs.}    & Evidence relevant to internal safety assessment; findings documented and traceable to intervention artifacts \\[2pt]
\textbf{Legal status}         & This mapping identifies evidence relevant to each provision; it does not establish legal compliance, which depends on the system, jurisdiction, and authoritative legal assessment \\
\end{tabular}

\noindent\rule{\linewidth}{0.4pt}
{\small\itshape End of Transition Report. This report was written using the \acr{} framework.}
\noindent\rule{\linewidth}{0.4pt}

\end{document}

%% file: defines.tex
\usepackage{xspace}
\usepackage{amsmath, amssymb, amsfonts}
\usepackage{bm}
\usepackage{mathtools}

\newcommand{\Mpre}{\ensuremath{M_{\text{pre}}}\xspace}
\newcommand{\Mpost}{\ensuremath{M_{\text{post}}}\xspace}
\newcommand{\Mplacebo}{\ensuremath{M_{\text{placebo}}}\xspace}

\newcommand{\acr}{\ensuremath{\mathrm{XAI}_{\Delta}}}

\newcommand{\acrb}{\ensuremath{\mathbf{XAI}_{\Delta}}}



%% file: diagrams/xaiandcxai_NIPS_2.tex
\scalebox{0.8}{%
\begin{tikzpicture}[>=Stealth, thick, node distance=0.6cm and 0.45cm]

    \definecolor{typeModel}{RGB}{245, 245, 245}
    \definecolor{typeExplainer}{RGB}{230, 240, 255}
    \definecolor{typeExplanation}{RGB}{235, 250, 235}
    \definecolor{typeBehavior}{RGB}{255, 235, 235}
    \definecolor{typeIntervention}{RGB}{255, 253, 220}
    \definecolor{boxDelta}{RGB}{235, 243, 255}
    \definecolor{boxXAI}{RGB}{255, 225, 185} 

    \tikzset{
        base/.style={draw, rectangle, rounded corners=2pt, minimum width=1.6cm, minimum height=0.8cm, align=center, font=\footnotesize},
        model/.style={base, fill=typeModel},
        explainer/.style={base, fill=typeExplainer},
        explanation/.style={base, fill=typeExplanation},
        behavior/.style={base, fill=typeBehavior},
        interv/.style={base, fill=typeIntervention}
    }

    \node[interv]    (I)      at (0,0)      {$I$};
    \node[model]     (Mpre)   [left=of I]   {$M_{\text{pre}}$};
    \node[behavior] (Bpre)  [left=of Mpre] {$B(M_{\text{pre}})$};
    \node[model]     (Mpost) [right=of I]  {$M_{\text{post}}$};
    \node[behavior] (Bpost) [right=of Mpost]{$B(M_{\text{post}})$};

    \node[explainer] (PhiPre)  [below=of Mpre]  {$\Phi^{(b)}(M_{\text{pre}})$};
    \node[explainer] (PhiPost) [below=of Mpost] {$\Phi^{(b)}(M_{\text{post}})$};

    \node[explainer]   (PhiDelta) [below=of I, yshift=-1.45cm]    {$\Phi^{(b)}_{\Delta}$};
    
    \node[explanation] (EPre)     [below=of PhiPre]  {$e^{(b)}_{\text{pre}}$};
    \node[explanation] (EPost)    [below=of PhiPost] {$e^{(b)}_{\text{post}}$};
    \node[explanation] (EDelta)   [below=0.8cm of PhiDelta] {$e^{(b)}_{\Delta}$};

    \begin{scope}[on background layer]
        \node[draw, fill=white, rounded corners=4pt, inner sep=9pt,
              fit=(Bpre) (Bpost) (Mpost) (EDelta)] (deltabox) {};

        \node[anchor=south west, font=\small\bfseries, yshift=2pt,xshift=100pt] 
              at (deltabox.north west) {The \acrb{} Framework};
        
        \node[draw, dotted, fill=boxXAI, rounded corners=4pt, inner sep=6pt,
              fit=(Mpost) (Bpost) (EPost),
              label={[anchor=north, font=\small\bfseries, xshift = 48pt, yshift=-100pt]XAI}] (xaibox_post) {};

        \node[draw, dotted, fill=boxXAI, rounded corners=4pt, inner sep=6pt,
              fit=(Mpre) (Bpre) (EPre),
              label={[anchor=north, font=\small\bfseries, xshift = -48pt, yshift=-100pt]XAI}] (xaibox_pre) {};
    \end{scope}

    \draw[->] (Mpre) -- (I);
    \draw[->] (I) -- (Mpost);
    \draw[->] (Mpre) -- (Bpre);
    \draw[->] (Mpost) -- (Bpost);
    \draw[->] (Mpre) -- (PhiPre);
    \draw[->] (Mpost) -- (PhiPost);
    \draw[->] (PhiPre) -- (EPre);
    \draw[->] (PhiPost) -- (EPost);
    
    \draw[->, blue] ([yshift=1.5pt,xshift=-0.5]Mpre.south east) -- (PhiDelta.north west);
   \draw[->, blue] ([yshift=1.5pt,xshift=0.5]Mpost.south west) -- (PhiDelta.north east);
    \draw[->, blue] (I) -- (PhiDelta.north);


    \draw[->,blue] (PhiDelta) -- (EDelta);

    \draw[dashed, ->, blue] (Bpre.south) |- (EDelta.west);
    \draw[dashed, ->, blue] (Bpost.south) |- (EDelta.east);
\end{tikzpicture}
}

%% file: tables/table2_new.tex
\begin{table*}[ht!]
\centering
\caption{\textbf{Families of comparative XAI (\acr{}) explainers for behavioral shift analysis.}
Each row describes a family of explanation protocols that extract \emph{paired evidence} from $(M_{\mathrm{pre}}, M_{\mathrm{post}})$ under matched conditions to produce a shift-focused explanation. Strengths and limitations indicate which explanatory claims about the behavioral shift $\Delta B$ are warranted by each family in isolation.}
\label{tab:method-zoo}
\scriptsize
\setlength{\tabcolsep}{2pt}
\renewcommand{\arraystretch}{1.08}

\begin{tabular}{p{1.85cm}p{3.75cm}p{3.75cm}p{3.75cm}}
\toprule
\textbf{$\Phi_{\Delta}^{(b)}$ family} &
\textbf{How to use} &
\textbf{Insights \& strengths} &
\textbf{Limitations} \\
\midrule

\textbf{Feature Attribution}
& Compute and contrast input-level attribution scores for $M_{\mathrm{pre}}$ and $M_{\mathrm{post}}$ under matched prompts, targets, and decoding. For prompting-based  interventions, compute attribution for the tokens in the intervention prompts. 
& Highlights how reliance on specific tokens or spans changes across checkpoints. Fast, local, and easily communicable; useful for hypothesis generation about input-linked drivers of $\Delta B$.
& Primarily correlational and method-sensitive; attribution differences depend on scoring and aggregation choices. They may miss internal changes not expressed as shifts in input reliance. \\[0.8ex]

\textbf{Probing}
& Train and compare probing classifiers to decode a target property from hidden states of multiple checkpoints. 
& Reveals representational changes and supports layer-wise localization of when a property becomes more linearly accessible.
& Requires predefined concepts and labels; probe success does not imply causal use. Sensitive to probe capacity and dataset construction. \\[0.8ex]

\textbf{Representation Similarity}
& Feed a shared input set through both models and compute layer- or component-wise similarity of activations or subspaces. Use similarity drops to nominate candidate regions of change.
& Provides a global divergence map useful for triage and narrowing the search space without labeled concepts.
& Quantifies difference but not semantics; similarity may be invariant to rotations or miss small but behaviorally critical changes. \\[0.8ex]

\textbf{Concept discovery}
& Extract interpretable directions or features from paired activations or weights and compare their presence or effects across checkpoints.
& Produces human-meaningful handles on behavioral change and enables downstream steering or projection-based tests.
& Concepts may be ill-defined or polysemantic; discovered directions may not generalize across layers or contexts and require careful interpretation. \\[0.8ex]

\textbf{Mechanistic interventions}
& Use $M_{\mathrm{pre}}$ as a reference for targeted interventions on $M_{\mathrm{post}}$ such as activation patching, ablations, or stitching.
& Provides the strongest mechanistic evidence, enabling falsifiable causal claims and fine-grained localization of responsible components.
& Labor-intensive and architecture-dependent; effects may be distributed and sensitive to intervention design. \\[0.8ex]

\textbf{Data attribution}
& For interventions like fine-tuning, attribute the observed behavioral difference to influential training examples, updates, or phases, producing ranked data- or update-level evidence.
& Connects $\Delta B$ to plausible training causes, supporting data-level remediation and auditability.
& Approximate and noisy at scale; credit assignment is difficult, especially for reinforcement learning or multi-stage training, and requires access to training artifacts. \\

\bottomrule
\end{tabular}
\end{table*}